\documentclass{article}

\usepackage[square,numbers]{natbib}

\usepackage[preprint]{neurips_2024} 

\usepackage[utf8]{inputenc} 
\usepackage[T1]{fontenc}    
\usepackage{hyperref}       
\usepackage{url}            
\usepackage{booktabs}       
\usepackage{amsfonts}       
\usepackage{nicefrac}       
\usepackage{microtype}      
\usepackage{xcolor}         

\usepackage{authblk}
\usepackage{amsmath}
\usepackage{mathtools}
\usepackage{algorithmic}
\usepackage{algorithm}
\usepackage{array}
\usepackage{subcaption}
\usepackage{graphicx}
\usepackage{textcomp}
\usepackage{float}
\usepackage{stfloats}
\usepackage{verbatim}
\usepackage{booktabs}
\usepackage{tabularx}
\usepackage{multirow}
\usepackage{makecell}
\usepackage{enumitem}
\usepackage{tablefootnote}
\usepackage{pdfcomment}
\usepackage{wrapfig}
\usepackage{pdfpages}

\newcolumntype{C}{>{\centering\arraybackslash}X}

\DeclareMathOperator*{\argmin}{\arg\!\min}

\begin{document}

\title{Neuronal Competition Groups with Supervised STDP for Spike-Based Classification}

\author[1]{Gaspard Goupy}
\author[1]{Pierre Tirilly}
\author[1,*]{Ioan Marius Bilasco}
\affil[1]{Univ. Lille, CNRS, Centrale Lille, UMR 9189 CRIStAL, F-59000 Lille, France}
\affil[*]{Corresponding author: marius.bilasco@univ-lille.fr}

\maketitle

\begin{abstract}
    Spike Timing-Dependent Plasticity (STDP) is a promising substitute to backpropagation for local training of Spiking Neural Networks (SNNs) on neuromorphic hardware.
    STDP allows SNNs to address classification tasks by combining unsupervised STDP for feature extraction and supervised STDP for classification.
    Unsupervised STDP is usually employed with Winner-Takes-All (WTA) competition to learn distinct patterns.
    However, WTA for supervised STDP classification faces unbalanced competition challenges.
    In this paper, we propose a method to effectively implement WTA competition in a spiking classification layer employing first-spike coding and supervised STDP training.
    We introduce the Neuronal Competition Group (NCG), an architecture that improves classification capabilities by promoting the learning of various patterns per class.
    An NCG is a group of neurons mapped to a specific class, implementing intra-class WTA and a novel competition regulation mechanism based on two-compartment thresholds.
    We incorporate our proposed architecture into spiking classification layers trained with state-of-the-art supervised STDP rules.
    On top of two different unsupervised feature extractors, we obtain significant accuracy improvements on image recognition datasets such as CIFAR-10 and CIFAR-100.
    We show that our competition regulation mechanism is crucial for ensuring balanced competition and improved class separation.
\end{abstract}

\section{Introduction} \label{sec:intro}
Neuromorphic computing~\citep{schumanSurveyNeuromorphicComputing2017} with Spiking Neural Networks (SNNs)~\citep{ponulakIntroductionSpikingNeural2011} is a promising solution to address the high energy consumption of Artificial Neural Networks (ANNs) on von Neumann architectures~\citep{zouBreakingNeumannBottleneck2021}.
However, direct training of SNNs on neuromorphic hardware faces a major constraint: implementing network-level communication is difficult and requires significant circuitry overhead~\citep{zenkeBrainInspiredLearningNeuromorphic2021}.
As a result, the learning mechanisms should be local, i.e., with weight updates based only on the activity of the two neurons that the synapse connects.

Training SNNs to achieve state-of-the-art (SOTA) performance is typically accomplished with adaptations of backpropagation (BP)~\citep{eshraghianTrainingSpikingNeural2021,dampfhofferBackpropagationBasedLearningTechniques2023}.
However, these methods are challenging to implement on neuromorphic hardware since they employ non-local learning~\citep{zenkeBrainInspiredLearningNeuromorphic2021,lillicrapBackpropagationBrain2020}.
In addition, they only rely on supervised learning, making them highly dependent on labeled data.
We believe that machine learning algorithms should minimize this dependence on supervision by leveraging unsupervised feature learning~\citep{bengioRepresentationLearningReview2013}.
Hence, an appealing classification system may comprise both unsupervised and supervised components, for feature extraction and classification, respectively.

Hebbian learning~\citep{hebbOrganizationBehavior1949} is an unsupervised and local alternative to BP, inspired by the principal form of plasticity observed in biological synapses.
Specifically, Spike Timing-Dependent Plasticity (STDP)~\citep{caporaleSpikeTimingDependent2008} is a form of Hebbian learning where the time difference between the input and output neuron spikes defines synaptic plasticity.
STDP could solve all the aforementioned limitations of BP, making it more suitable for on-chip training on neuromorphic hardware~\citep{saighiPlasticityMemristiveDevices2015,khacefSpikebasedLocalSynaptic2023}.
STDP is particularly effective with first-spike coding~\citep{falezImprovingSpikingNeural2019,guoNeuralCodingSpiking2021}, where neurons can fire at most once per sample.
Using one spike per neuron presents several advantages, including energy efficiency~\citep{rueckauerConversionAnalogSpiking2018,parkT2FSNNDeepSpiking2020}, fast information transfer~\citep{rullenRateCodingTemporal2001}, and high information capacity~\citep{augeSurveyEncodingTechniques2021}.
While primarily used for unsupervised feature learning~\citep{kheradpishehSTDPbasedSpikingDeep2018,falezMultiLayeredSpikingNeural2019,el-assal2D3DConvolutional2022}, STDP can be extented to supervised learning~\citep{mozafariFirstspikebasedVisualCategorization2018,zhaoGLSNNMultiLayerSpiking2020,liuSSTDPSupervisedSpike2021}.
As a result, SNNs can perform classification tasks by combining unsupervised STDP for feature extraction and supervised STDP for classification~\citep{shresthaStableSpiketimingDependent2017,mozafariBioinspiredDigitRecognition2019,thieleEventBasedTimescaleInvariant2018,leeDeepSpikingConvolutional2019,goupyPairedCompetingNeurons2024}.
Employing the same type of local learning rule for both feature extraction and classification ensures consistency and may facilitate hardware implementation.

Unsupervised STDP is commonly paired with Winner-Takes-All (WTA) competitive learning to promote the discovery of distinct patterns~\citep{diehlUnsupervisedLearningDigit2015,kheradpishehSTDPbasedSpikingDeep2018,ferreUnsupervisedFeatureLearning2018,falezMultiLayeredSpikingNeural2019}.
In a WTA framework with first-spike coding, lateral inhibition is implemented to ensure that only the first neuron to fire receives a weight update.
In addition, homeostatic mechanisms, such as threshold adaptation, must be employed to regulate the competition among neurons~\citep{leeTrainingDeepSpiking2016,falezMultiLayeredSpikingNeural2019,quEfficientHardwareFriendlyMethods2020,haoBiologicallyPlausibleSupervised2020}.
For supervised STDP, WTA competition is also appealing as it may improve the learning capabilities of a classification layer with multiple neurons per class~\citep{mozafariFirstspikebasedVisualCategorization2018}.
Specifically, intra-class WTA can promote the learning of various class-specific patterns.
However, supervised STDP classification with WTA competition has been poorly studied in the literature~\citep{mozafariFirstspikebasedVisualCategorization2018,bethiOptimizedDeepSpiking2022} and presents unbalanced competition challenges.
Indeed, there is a lack of competition regulation methods, and regular threshold adaptation rules can lead to unfair decision-making since output neurons may use different thresholds for inference.

In this paper, we address WTA-based competitive learning in supervised STDP.
We aim to implement effective WTA competition in a spiking classification layer employing first-spike coding and SOTA supervised STDP rules.
Our main contributions can be summarized as follows:
\begin{enumerate}[leftmargin=*]
    \item We introduce the Neuronal Competition Group (NCG), an architecture that improves classification capabilities by promoting the learning of various patterns per class. In the classification layer, each class is mapped to an NCG: a group of neurons using intra-class WTA and competition regulation.
    \item To ensure both balanced intra-class competition and fair decision-making, we design a competition regulation mechanism based on two-compartment thresholds. Neurons are equipped with a fixed threshold for decision-making, along with an adaptive threshold used to regulate the frequency at which they update their weights on samples of their class.
    \item To validate our architecture with input features of varying quality, we incorporate NCGs into spiking classification layers placed on top of two Hebbian-based feature extractors. Using NCGs with SOTA supervised STDP rules, we obtain significant accuracy gains on image recognition datasets: MNIST, Fashion-MNIST, CIFAR-10, and CIFAR-100. We show that our competition regulation mechanism is crucial for ensuring balanced competition and improved class separation.
\end{enumerate}
The source code is publicly available at: \url{https://gitlab.univ-lille.fr/fox/snn-ncg}.

\section{Related Work} \label{sec:related}

\paragraph{Supervised Training with STDP}
Supervised training of SNNs with STDP introduces an error signal~\citep{fremauxNeuromodulatedSpikeTimingDependentPlasticity2015} that is used to guide the STDP updates.
Several supervised adaptations of STDP are reported in the literature~\citep{ponulakSupervisedLearningSpiking2010,shresthaStableSpiketimingDependent2017,tavanaeiBPSTDPApproximatingBackpropagation2019,shresthaApproximatingBackPropagationBiologically2019,haoBiologicallyPlausibleSupervised2020,zhaoGLSNNMultiLayerSpiking2020,saraniradAssemblybasedSTDPNew2022}.
Yet, the aforementioned rules are designed to train SNNs with multiple spikes per neuron, which is not as efficient as first-spike coding.
The literature exploring supervised STDP training of SNNs with one spike per neuron is limited~\citep{mozafariFirstspikebasedVisualCategorization2018,liuSSTDPSupervisedSpike2021,goupyPairedCompetingNeurons2024}.
Reward-Modulated STDP (R-STDP)~\citep{mozafariFirstspikebasedVisualCategorization2018} involves supervised training by adjusting the sign of STDP.
The employed error is fairly simple (\(+1\) or \(-1\)), resulting in inaccurate weight updates.
SSTDP~\citep{liuSSTDPSupervisedSpike2021} and S2-STDP~\citep{goupyPairedCompetingNeurons2024} are more recent methods that compute temporal errors to adjust both the sign and the intensity of weight updates, making them more accurate.
However, unlike R-STDP, these methods cannot be used with various neurons per class in a classification layer.

\paragraph{Competitive Learning for Classification}
Employing groups of neurons is an effective approach for improving the learning capabilities of a classification layer~\citep{beyelerCategorizationDecisionmakingNeurobiologically2013,luoFirstErrorBasedSupervised2019,wangCompSNNLightweightSpiking2021}.
To maximize knowledge within the layer and learn distinct patterns, WTA-based competitive learning can be employed~\citep{ferreUnsupervisedFeatureLearning2018}.
While WTA competition is widely adopted in unsupervised learning~\citep{diehlUnsupervisedLearningDigit2015,kheradpishehSTDPbasedSpikingDeep2018,ferreUnsupervisedFeatureLearning2018,falezMultiLayeredSpikingNeural2019}, its application to supervised learning is limited.
Prior work~\citep{leeTrainingDeepSpiking2016,kulkarniSpikingNeuralNetworks2018,lobovCompetitiveLearningSpiking2020} implemented WTA competition at the classification layer but only with one neuron per class, making it impossible to learn various class-specific patterns.
Conversely, reward-based approaches~\citep{mozafariFirstspikebasedVisualCategorization2018,bethiOptimizedDeepSpiking2022}, such as R-STDP, are the only methods that implement WTA with multiple neurons per class.
Through lateral inhibition, neurons compete for weight updates, both within the same class (intra-class WTA) and across different classes (inter-class WTA).
Intra-class WTA enables neurons to learn patterns from distinct samples.
However, inter-class WTA prevents accurate control over the time difference between the spikes of target and non-target neurons (i.e. neurons mapped or not to the class), as only one neuron is updated per sample.
In~\citep{goupyPairedCompetingNeurons2024}, solely intra-class WTA and two neurons per class were employed to promote specialization toward target and non-target samples.
Nonetheless, to the best of our knowledge, no prior work solely employed intra-class WTA to promote the learning of various class-specific patterns.

\paragraph{Competition Regulation}
In WTA-based competitive learning, it is crucial to implement regulation (also called homeostatic) mechanisms to ensure balanced competition among neurons~\citep{ferreUnsupervisedFeatureLearning2018,falezMultiLayeredSpikingNeural2019,quEfficientHardwareFriendlyMethods2020}.
A simple solution is to use dropout~\citep{srivastavaDropoutSimpleWay2014} on the output neurons, as done with R-STDP~\citep{mozafariFirstspikebasedVisualCategorization2018}, where some neurons of each class are randomly deactivated during training to encourage weight updates on distinct samples.
Yet, this solution is not optimal due to its stochastic nature.
Other regulation mechanisms involve threshold adaptation~\citep{leeTrainingDeepSpiking2016, falezMultiLayeredSpikingNeural2019,haoBiologicallyPlausibleSupervised2020}, by increasing or reducing thresholds to promote or discourage firing.
While threshold adaptation is an effective solution to ensure balanced competition, using different thresholds across neurons may prevent fair decision-making since their firing time is tied to their thresholds.
Prior work employed multiple thresholds per neuron~\citep{thieleEventBasedTimescaleInvariant2018,yuConstructingAccurateEfficient2022} but the authors did not incorporate threshold adaptation mechanisms.
In this work, we draw inspiration from multi-thresholds and threshold adaptation to design a competition regulation mechanism based on two-compartment thresholds, ensuring both balanced competition and fair decision-making.

\section{Preliminaries} \label{sec:prelim}

\subsection{Neuron Model}
To align with first-spike coding, we use the Single-Spike Integrate-and-Fire (SSIF) model~\citep{goupyUnsupervisedEfficientLearning2023}, where neurons can fire at most once per sample.
Since each neuron emits a single spike, the intensity of its activation is encoded via a firing timestamp: the most activated neuron fires first.
The membrane potential \(V_j\) of a neuron $n_j$ is expressed as:
\begin{equation}
    \begin{aligned}
        \frac{\partial V_j\left(t\right)}{\partial t} & = \sum_i W_{ij} \cdot S_i\left(t\right) \\
        S_i\left(t\right)                             & = \begin{cases}
            1 & \text{if } V_i\left(t\right) \geq \theta \\
            0 & \text{o.w. }
        \end{cases},
    \end{aligned}
    \label{eq:if-model}
\end{equation}
where \(t\) is the timestamp, \(S_i\left(t\right)\) indicates the presence or absence of a spike from input neuron \(n_i\) at timestamp \(t\), and \(W_{ij}\) is the weight of the synapse from \(n_i\) to \(n_j\).
When the membrane potential of a neuron reaches its firing threshold \(\theta\), the neuron emits a spike, resets its membrane potential to zero, and remains deactivated until the next sample is shown.
In our simulations, firing timestamps are represented by floating-point values to align with event-driven neuromorphic hardware.

\subsection{Spiking Classification Layer}
The spiking classification layer is a fully-connected architecture comprising, for a \(C\)-class problem, \(N = C \times M\) neurons \(\left(n_1, \ldots, n_N\right)\), where \(M\) is the number of neurons per class.
Each neuron \(n_j\) is mapped to a class \(c_j\).
Aligned with the SSIF model, we employ first-spike-based decision-making: the first output neuron to fire predicts the class.
This method removes the need to propagate the entire input for inference, which can reduce computation time and the number of generated spikes.
Formally, the prediction \(\hat{y}\) of the SNN is defined as:
\begin{equation}
    \begin{aligned}
        \hat{y} & = c_{j^*}                                           \\
        j^*     & = \argmin_{j \in \left[1,N\right]}\left(t_j\right),
    \end{aligned}
\end{equation}
where \(t_j\) denotes the firing timestamp of neuron \(n_j\).
If multiple neurons fire at the same timestamp, the one with the highest membrane potential is selected.
In practice, the method employed to select a neuron in the event of a tie has little effect on performance.
In this work, the classification layer is placed on top of an unsupervised feature extraction network.

\subsection{Supervised STDP Training}
Neurons of the classification layer are trained with a supervised STDP rule.
At the end of a sample presentation, weights of non-inhibited neurons are updated with an error-modulated additive STDP:
\begin{equation}
    \Delta W_{ij}=
    \begin{cases}
        e_j \times A^{+} & \text{if } t_j \geq t_i \\
        e_j \times A^{-} & \text{o.w. }
    \end{cases},
    \label{eq:s2stdp-update}
\end{equation}
where \(\Delta W_{ij}\) is the weight change (such as  \(W_{ij} \vcentcolon = W_{ij} + \Delta W_{ij}\)), \(e_j\) is the error of neuron \(n_j\), \(A^+ > 0\) and \(A^- < 0\) are the learning rates.
These two learning rates control learning speed and determine the relative importance of long-term potentiation (\(A^+\)) versus long-term depression (\(A^-\)) in the learning process.
Weights are manually clipped in \(\left[w_\mathrm{min}, w_\mathrm{max}\right]\) after each update to ensure that they remain within a controlled range.

\paragraph{R-STDP}
Reward-Modulated STDP (R-STDP)~\citep{mozafariFirstspikebasedVisualCategorization2018} is a rule combined with WTA competition.
For each sample, only the first neuron to fire receives a weight update.
The error is \(e_j=+1\) if \(n_j\) is mapped to the class of the sample, \(e_j=-1\) otherwise.
In practice, a classification layer trained with R-STDP requires multiple neurons per class to achieve reasonable performance.
R-STDP is usually employed in conjunction with adaptive learning rates to reduce overfitting, and dropout to facilitate the learning of various patterns per class~\citep{mozafariFirstspikebasedVisualCategorization2018}.

\paragraph{SSTDP}
Supervised STDP (SSTDP)~\citep{liuSSTDPSupervisedSpike2021} is a rule with SOTA performance.
It is employed with one neuron per class and without WTA.
This rule provides high adaptability to input data by dynamically computing temporal errors for each sample, based on the average firing time \(\overline{T}\) in the layer:
\begin{equation}
    e_j= t_j - \begin{cases}
        \min \left\{ t_j, \overline{T}-\frac{C-1}{C}g \right\} & \text{if } c_j = y    \\
        \max \left\{ t_j, \overline{T}+\frac{1}{C}g \right\}   & \text{if } c_j \neq y
    \end{cases},
    \label{eq:sstdp-error}
\end{equation}
where \(y\) is the class of the sample, and \(g\) is a hyperparameter that controls the desired distance from \(\overline{T}\).
The optimal value of \(g\) partly depends on the input spike distribution: a narrower distribution requires a smaller \(g\).
For each sample, due to the $\min$ and $\max$ functions, only the target neuron firing after \(\overline{T}-\frac{C-1}{C}g\) and the non-target neurons firing before \(\overline{T}+\frac{1}{C}g\) update their weights.

\paragraph{S2-STDP}
Stabilized Supervised STDP (S2-STDP)~\citep{goupyPairedCompetingNeurons2024} addresses two limitations of SSTDP: the limited number of updates per epoch and the saturation of firing timestamps toward the maximum firing time.
In this rule, neurons are trained to fire at desired timestamps instead of time ranges:
\begin{equation}
    e_j= t_j - \begin{cases}
        \overline{T}-\frac{C-1}{C}g & \text{if } c_j = y    \\
        \overline{T}+\frac{1}{C}g   & \text{if } c_j \neq y
    \end{cases}.
    \label{eq:s2stdp-error}
\end{equation}
This enables more accurate control over the output firing times and reduces the saturation effect.
Also, weight normalization is used to keep a similar weight average across neurons during learning~\citep{goupyPairedCompetingNeurons2024}.

\section{Methods} \label{sec:method}

\subsection{Neuronal Competition Group} \label{sec:ncg}
Training a classification layer involves teaching neurons to recognize a pattern specific to their class from the input samples.
Different samples from a given class can contain distinct, mutually exclusive patterns or combinations of patterns.
Learning all these patterns concurrently with one neuron can be challenging and impose strong generalization constraints on its weights, especially when using a single supervised layer.
Employing multiple neurons per class to learn various class-specific patterns may reduce these constraints and enable the emergence of more specialized patterns that better represent the training set distribution.
Building on this concept, we introduce the Neuronal Competition Group (NCG), an architecture promoting the learning of various class-specific patterns through intra-class WTA and competition regulation.

\begin{figure}[t]
    \centering
    \includegraphics[width=1\linewidth]{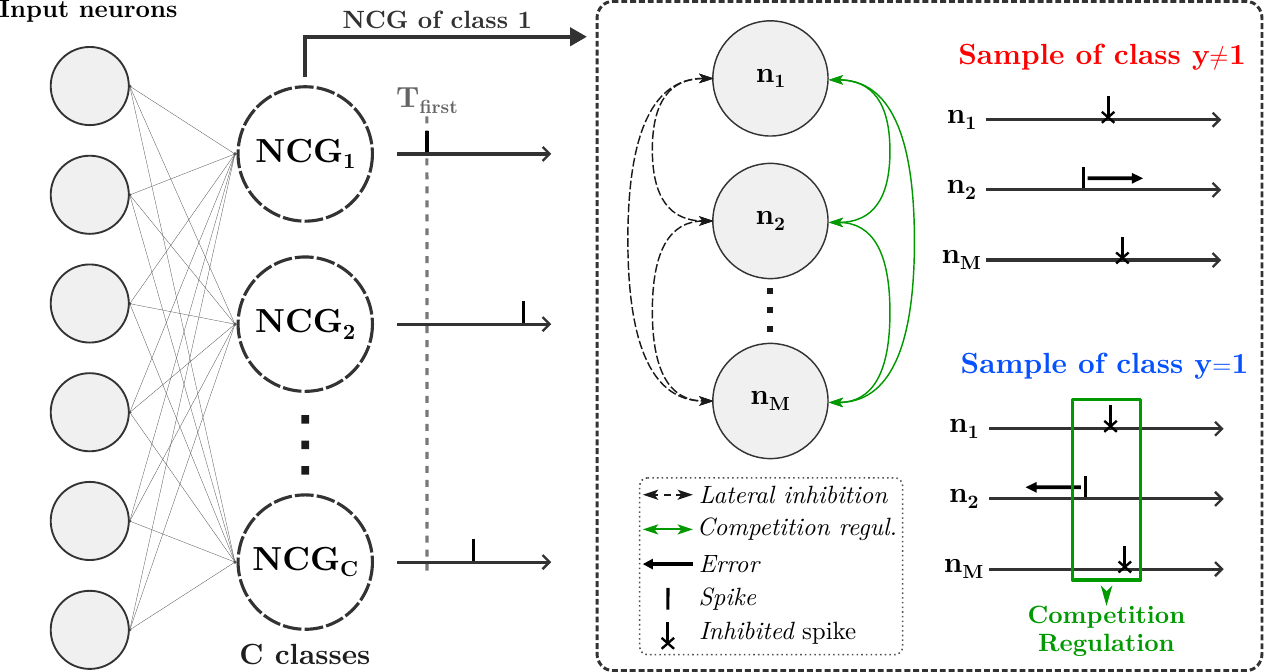}
    \caption{Spiking classification layer with Neuronal Competition Groups (NCGs). In this layer, each class is mapped to an NCG and the prediction is based on the first spike. An NCG is a group of \(M\) neurons connected with lateral inhibition to enable intra-class WTA competition: the first neuron to fire inhibits the other ones and undergoes a weight update based on a temporal error (which depends on the learning rule considered). The sign and amplitude of the error pushes neurons to fire earlier (positive sign) or later (negative sign). Competition regulation occurs only within the NCG mapped to the class of the input sample to ensure balanced competition among neurons on samples of their class. NCGs improve the classification capabilities of a layer by promoting the learning of various patterns per class.}
    \label{fig:ncg}
\end{figure}
The NCG architecture, illustrated in Figure~\ref{fig:ncg}, augments a classification layer by mapping each class to an NCG instead of independent neurons.
An NCG is a group of \(M\) neurons that aim to learn different patterns from samples of their mapped class.
Neurons of an NCG are interconnected with lateral inhibition, such as, for a given sample, the first neuron to fire within a group emits an inhibitory signal that prevents the other ones from firing.
Lateral inhibition induces competitive learning through intra-class WTA: only the first neuron to fire undergoes the weight update.
There is no lateral inhibition between NCGs (i.e. inter-class WTA).
Hence, each sample triggers exactly one weight update per NCG.
Removing inter-class WTA enables more accurate control over the time difference between the spikes of target and non-target neurons, which can improve class separation~\citep{goupyPairedCompetingNeurons2024}.
Each time a sample is presented during training, competition regulation is triggered in the NCG mapped to the class of the sample.
This mechanism ensures balanced competition within the NCGs, which facilitate the learning of various class-specific patterns.

\subsection{Competition Regulation} \label{sec:cr}
Preliminary experiments highlighted that intra-class WTA competition provided by lateral inhibition is not enough to ensure balanced competition.
In practice, for each NCG, one neuron tends to dominate the others, receiving the majority of the weight updates from samples of its class.
Although threshold adaptation can be employed to regulate competition~\citep{leeTrainingDeepSpiking2016,falezMultiLayeredSpikingNeural2019}, in a decision-making context, different thresholds between neurons may lead to unfair decisions because predictions are based on the first spike.
To ensure both balanced intra-class competition and fair decision-making, we introduce a competition regulation mechanism based on two-compartment thresholds.

In the classification layer, all the neurons are equipped with an identical and fixed threshold, denoted as the test threshold \(\theta\).
This threshold remains fixed to ensure fair decision-making during inference, as the class is predicted by the neuron that fires first.
On top of that, neurons are equipped with an additional varying threshold, denoted as the training threshold \(\theta'\).
Neurons switch to their \(\theta'\) only when they are exposed to samples of their class during training.
Otherwise, they always employ \(\theta\), both for inference and for samples of other classes during training.
\(\theta'\) is the key component to balance intra-class competition: it can be increased or decreased to encourage or reduce neuron firing on samples of its class.
Each time a neuron receives a weight update from a sample of its class, competition regulation is triggered across neurons of its NCG.
Their \(\theta'\) are updated as follows:
\begin{equation}
    \begin{aligned}
        \Delta \theta'_j & =
        \begin{cases}
            +\eta_{\mathrm{th}} \cdot \frac{M-1}{M} & \text {if } t_j=\min \left\{t_1, \cdots, t_{M}\right\} \\
            -\eta_{\mathrm{th}} \cdot \frac{1}{M}   & \text {o.w. }
        \end{cases}
        \\
        \theta'_j        & \vcentcolon =\max \left\{\theta_j, \theta'_j + \Delta \theta'_j\right\},
    \end{aligned}
    \label{eq:cr}
\end{equation}
where \(\theta'_j\) and \(\theta_j\) are the training and test thresholds of neuron \(n_j\), \(M\) is the number of neurons in the NCG, \(\eta_{\mathrm{th}}\) is the threshold learning rate, and \(t_j\) is the firing timestamp of neuron \(n_j\).
If several neurons fire at the same timestamp, the one with the highest membrane potential is selected (this has no impact on performance).
\(\theta'\) is reset to \(\theta\) between epochs and its minimum achievable value is \(\theta\).
These two components ensure that neurons learn patterns consistent with \(\theta\), which is the threshold that they use for inference.
\(\eta_{\mathrm{th}}\) defines the strength of competition regulation: higher values favor more balanced competition but may deteriorate pattern learning since \(\theta'\) tend to increase within an epoch. It should be chosen together with the initial threshold (a higher threshold may require a higher \(\eta_{\mathrm{th}}\)).
To achieve better convergence and robustness, an annealing factor \(\beta_{\mathrm{th}}\) can be added to reduce \(\eta_{\mathrm{th}}\) after each epoch, such as \(\eta_{\mathrm{th}} \vcentcolon = \eta_{\mathrm{th}} \cdot \beta_{\mathrm{th}}\).
\(\beta_{\mathrm{th}}\) affects the number of epochs during which competition regulation occurs and should be adjusted according to \(\eta_{\mathrm{th}}\): higher \(\eta_{\mathrm{th}}\) requires lower \(\beta_{\mathrm{th}}\).

\subsection{Neuron Labeling} \label{sec:nt}
In~\citep{goupyPairedCompetingNeurons2024}, WTA competition enhances a classification layer with two neurons per class and S2-STDP training by naturally promoting, for each class, neuron specialization toward target or non-target samples.
This behavior can also be implemented with NCGs, but it requires explicit neuron labeling to ensure that all neurons but one specialize toward samples of their class.
In such cases, one neuron within each NCG can be labeled as non-target, whereas the others can be labeled as target.
All the neurons are connected with lateral inhibition but only target neurons are connected with competition regulation.
Hence, if the non-target neuron fires first for a sample of the class, it prevents target neurons from updating their weights and applying competition regulation.
Regardless of the class of the sample, a non-target neuron \(n_j\) winning the competition always updates its weights as if \(c_j \neq y\) in Equation~\ref{eq:s2stdp-error}.
STDP training remains unchanged for target neurons.
However, in Equation~\ref{eq:cr}, \(M\) must be updated as it refers to the number of target neurons.
In Supplementary Material (Section~1), we provide the overall algorithm for training a spiking classification layer with our proposed methods.

\section{Experiments} \label{sec:exp}

\subsection{Experimental Setup} \label{sec:setup}

\subsubsection{Datasets} \label{sec:setup-dataset}
We select four image recognition datasets of growing complexity: MNIST~\citep{lecunGradientbasedLearningApplied1998}, Fashion-MNIST~\citep{xiaoFashionMNISTNovelImage2017}, CIFAR-10~\citep{krizhevskyLearningMultipleLayers2009}, and CIFAR-100~\citep{krizhevskyLearningMultipleLayers2009}.
MNIST and Fashion-MNIST comprise \(28\times28\) grayscale images, \(60,000\) samples for training and \(10,000\) for testing, categorized into \(10\) classes.
CIFAR-10 and CIFAR-100 comprise \(32 \times 32\) RGB images, \(50,000\) for training and \(10,000\) for testing.
They consist of, respectively, \(10\) and \(100\) classes.

\subsubsection{Classification Pipeline} \label{sec:setup-pipeline}
Our classification system consist of a feature extractor trained with unsupervised Hebbian-based learning, followed by a spiking classification layer trained with supervised STDP.
Training is layer-wise: the feature extractor is trained entirely before the training of the classification layer starts.
The complete pipeline of our classification system is illustrated in Supplementary Material (Section~2.1).

\subsubsection{Unsupervised Feature Extractors} \label{sec:setup-extractor}
To improve image representation before classification without labeled data, we consider two Hebbian-based unsupervised feature extractors built on Convolutional Neural Networks (CNNs):
\begin{enumerate}
    \item STDP-CSNN~\citep{falezMultiLayeredSpikingNeural2019}, a single-layer spiking CNN trained with STDP;
    \item SoftHebb-CNN~\citep{journeHebbianDeepLearning2023}, a three-layer non-spiking CNN trained with SoftHebb.
\end{enumerate}
Employing various feature extractors allows us to validate our methods with input features of varying quality.
These two feature extractors are SOTA in their category (spiking/non-spiking), share local learning properties, and offer different baseline performances.
In particular, SoftHebb-CNN, while not spike-based, is a relevant alternative for exploring classification using features provided by effective multi-layer local learning.
The extracted feature maps are flattened to match the fully-connected architecture of the classification layer.
Since SoftHebb-CNN is not spike-based, its output features are encoded into spike timestamps with a form of first-spike coding~\citep{thorpeSpikebasedStrategiesRapid2001}.
STDP-CSNN outputs \(4,608\) features for MNIST/Fashion-MNIST, and \(6,272\) for CIFAR-10/100.
SoftHebb-CNN outputs \(13,824\) features for MNIST/Fashion-MNIST, and \(24,576\) for CIFAR-10/100.
Aligned with first-spike coding, each feature is a single floating-point spike timestamp in \(\left[0, 1\right]\).
Additional details are reported in Supplementary Material (Section~2.2).

\subsubsection{Spiking Classification Layers} \label{sec:setup-classification-layer}
We train fully-connected spiking classification layers with three existing supervised STDP rules designed for one spike per neuron: R-STDP~\citep{mozafariFirstspikebasedVisualCategorization2018}, SSTDP~\citep{liuSSTDPSupervisedSpike2021}, and S2-STDP~\citep{goupyPairedCompetingNeurons2024}.
We incorporate the NCG architecture into classification layers trained with SSTDP and S2-STDP, denoted as SSTDP+NCG and S2-STDP+NCG, respectively.
R-STDP is incompatible with NCGs since it requires inter-class WTA for weight convergence.
Unless otherwise specified, we set \(M=5\) neurons per class for NCG-based methods, which is the smallest value providing, on average, near-optimal performance on the evaluated datasets (see Section~3.2 of Supplementary Material).
We evaluated R-STDP with both \(M=5\) and \(M=20\), the value providing near-optimal performance for this rule.
With S2-STDP+NCG, one neuron of each NCG is labeled as non-target, as detailed in Section~\ref{sec:nt}.

\subsubsection{Protocol} \label{sec:setup-protocol}
We divide our experimental protocol into two phases: hyperparameter optimization and evaluation.
In both phases, we employ an early stopping mechanism (with a patience \(\rho\)) during training to prevent overfitting.
For hyperparameter optimization, we construct a validation set from the training set by randomly selecting, for each class, a percentage \(\nu\) of its samples.
Then, we use the gridsearch algorithm to optimize the hyperparameters of the spiking classification layer (for each rule, dataset, and feature extractor).
No gridsearch is performed on CIFAR-100: we employ the optimized hyperparameters from CIFAR-10, given the similarities between the two datasets.
Additional details regarding hyperparameters are provided in Supplementary Material (Section~2.3).
For evaluation, we employ the K-fold cross-validation strategy.
We divide the training set into \(K\) subsets and train \(K\) models, each using a different subset for validation while the remaining \(K-1\) subsets are used for training.
Each model is trained with a different seed.
Then, we evaluate the trained models on the test set and we compute the mean test accuracy and standard deviation (1-sigma).
We use \(\rho=10\), \(K=10\) and \(\nu=\frac{1}{K}\) (i.e. we allocate \(10\)\% of the training sets for validation).

\subsection{Accuracy Comparison}
\begin{table}[t]
    \centering
    \setlength{\extrarowheight}{2pt}
    \footnotesize
    \caption{Accuracy of spiking classification layers trained with STDP-based methods, on top of Hebbian-based unsupervised feature extractors.}
    \begin{tabularx}{\linewidth}{@{}Ccccc@{}}
        \toprule
        \multirow{2}{*}{Dataset}       & \multirow{2}{*}{Method}     & \multirow{2}{*}{\makecell{Neurons                                                                               \\ per class}}   & \multicolumn{2}{c}{Accuracy (Mean\(\pm\)Std \%)} \\
                                       &                             &                                   & STDP-CSNN                            & SoftHebb-CNN                         \\
        \midrule
        \multirow{6}{*}{MNIST}         & \multirow{2}{*}{R-STDP}     & 5                                 & \(96.82 \pm 0.29\)                   & \(97.72 \pm 0.26\)                   \\
                                       &                             & 20                                & \(97.49 \pm 0.12\)                   & \(98.24 \pm 0.15\)                   \\
        \cmidrule(lr){2-5}
                                       & SSTDP                       & 1                                 & \(96.44 \pm 0.09\)                   & \(98.52 \pm 0.16\)                   \\
                                       & SSTDP+NCG \textit{(ours)}   & 5                                 & \(97.30 \pm 0.09\)                   & \(98.96 \pm 0.06\)                   \\
        \cmidrule(lr){2-5}
                                       & S2-STDP                     & 1                                 & \(97.74 \pm 0.06\)                   & \(98.81 \pm 0.09\)                   \\
                                       & S2-STDP+NCG \textit{(ours)} & 5                                 & \textbf{98.92} \(\pm\) \textbf{0.07} & \textbf{99.17} \(\pm\) \textbf{0.07} \\
        \midrule
        \multirow{6}{*}{Fashion-MNIST} & \multirow{2}{*}{R-STDP}     & 5                                 & \(78.40 \pm 0.89\)                   & \(87.32 \pm 0.76\)                   \\
                                       &                             & 20                                & \(82.17 \pm 0.38\)                   & \(88.06 \pm 0.29\)                   \\
        \cmidrule(lr){2-5}
                                       & SSTDP                       & 1                                 & \(85.26 \pm 0.17\)                   & \(89.36 \pm 0.24\)                   \\
                                       & SSTDP+NCG \textit{(ours)}   & 5                                 & \(87.59 \pm 0.11\)                   & \(91.06 \pm 0.10\)                   \\
        \cmidrule(lr){2-5}
                                       & S2-STDP                     & 1                                 & \(85.89 \pm 0.27\)                   & \(90.61 \pm 0.19\)                   \\
                                       & S2-STDP+NCG \textit{(ours)} & 5                                 & \textbf{88.72} \(\pm\) \textbf{0.23} & \textbf{91.86} \(\pm\) \textbf{0.14} \\
        \midrule
        \multirow{6}{*}{CIFAR-10}      & \multirow{2}{*}{R-STDP}     & 5                                 & \(62.12 \pm 0.62\)                   & \(74.12 \pm 0.34\)                   \\
                                       &                             & 20                                & \(65.92 \pm 0.68\)                   & \(75.54 \pm 0.57\)                   \\
        \cmidrule(lr){2-5}
                                       & SSTDP                       & 1                                 & \(60.87 \pm 0.53\)                   & \(76.57 \pm 0.58\)                   \\
                                       & SSTDP+NCG \textit{(ours)}   & 5                                 & \(64.05 \pm 0.48\)                   & \(78.53 \pm 0.32\)                   \\
        \cmidrule(lr){2-5}
                                       & S2-STDP                     & 1                                 & \(61.08 \pm 0.17\)                   & \(76.90 \pm 0.27\)                   \\
                                       & S2-STDP+NCG \textit{(ours)} & 5                                 & \textbf{66.41} \(\pm\) \textbf{0.17} & \textbf{79.55} \(\pm\) \textbf{0.23} \\
        \midrule
        \multirow{6}{*}{CIFAR-100}     & \multirow{2}{*}{R-STDP}     & 5                                 & \(32.07 \pm 0.38\)                   & \(48.27 \pm 0.36\)                   \\
                                       &                             & 20                                & \(34.77 \pm 0.44\)                   & \(49.25 \pm 0.48\)                   \\
        \cmidrule(lr){2-5}
                                       & SSTDP                       & 1                                 & \(28.49 \pm 0.49\)                   & \(48.73 \pm 0.39\)                   \\
                                       & SSTDP+NCG \textit{(ours)}   & 5                                 & \(31.19 \pm 0.27\)                   & \(49.81 \pm 0.23\)                   \\
        \cmidrule(lr){2-5}
                                       & S2-STDP                     & 1                                 & \(29.39 \pm 0.19\)                   & \(49.17 \pm 0.29\)                   \\
                                       & S2-STDP+NCG \textit{(ours)} & 5                                 & \textbf{35.90} \(\pm\) \textbf{0.42} & \textbf{53.49} \(\pm\) \textbf{0.18} \\
        \bottomrule
    \end{tabularx}
    \label{tab:acc-comp}
\end{table}
We compare, in Table~\ref{tab:acc-comp}, the performance of the different STDP-based methods for training a spiking classification layer (see Section~\ref{sec:setup-classification-layer}) placed on top of each unsupervised feature extractor (see Section~\ref{sec:setup-extractor}).
Our proposed NCG architecture consistently improves the performance of SSTDP and S2-STDP across all datasets and feature extractors.
The accuracy improvement tends to scale with the complexity of the dataset.
With S2-STDP and the STDP-CSNN feature extractor, we measure an increase of \(1.18\)~pp on MNIST, \(2.83\)~pp on Fashion-MNIST, \(5.33\)~pp on CIFAR-10, and \(6.51\)~pp on CIFAR-100.
S2-STDP always outperforms SSTDP and enables higher accuracy improvement when paired with NCG as it leverages neuron labeling.
While S2-STDP surpasses R-STDP when the input features are well-captured, it falls behind in scenarios involving lower-quality features (CIFAR-10 with STDP-CSNN, CIFAR-100), as R-STDP can learn various patterns per class.
S2-STDP+NCG effectively bridges this gap, outperforming R-STDP on both simpler and harder tasks while requiring four times fewer neurons per class.
When R-STDP is used with the same number of neurons as S2-STDP+NCG, the accuracy gap is even larger.
These results highlight that WTA-based competitive learning in a supervised context can be achieved without inter-class WTA.
Employing solely intra-class WTA and accurate STDP updates enables more effective training.

Regarding the literature on SNNs with fully-supervised local-based learning, SOTA performance is achieved by STiDi-BP (one spike per neuron)~\citep{mirsadeghiSpikeTimeDisplacementbased2023} on MNIST (\(99.20\)\% with a 3-layer SNN) as well as Fashion-MNIST (\(92.80\)\% with a 4-layer SNN), and by EMSTDP (multiple spikes per neuron)~\citep{shresthaInHardwareLearningMultilayer2021} on CIFAR-10 (\(64.40\)\% with a 4-layer SNN).
We did not find any work reporting results on CIFAR-100.
For approaches combining unsupervised and supervised local learning, SOTA performance is achieved by R-STDP~\citep{mozafariBioinspiredDigitRecognition2019} on MNIST (\(97.20\)\% with a 3-layer SNN) and by Sym-STDP~\citep{haoBiologicallyPlausibleSupervised2020} on Fashion-MNIST (\(85.31\)\% with a 2-layer SNN).
We did not find any work reporting results on CIFAR-10/100.
Our best models, comprising 4-layer networks with only one supervised layer, achieve \(99.17\)\% on MNIST, \(91.86\)\% on Fashion-MNIST, and \(79.55\)\% on CIFAR-10.
Our results closely match or surpass fully-supervised SOTA work and outperform semi-supervised SOTA work.
Yet, it is important to acknowledge the role of the feature extractor in the final performance.
There remains a huge gap between local-based and global-based approaches in terms of accuracy. In Supplementary Material (Section~4), we compare our methods with global-based approaches to highlight that, despite the accuracy gap, local-based methods show greater computational efficiency, lower memory usage, reduced energy consumption, and easier hardware implementation, justifying further exploration.

\subsection{Ablation Study} \label{sec:ablation}
\begin{table}[ht]
    \caption{Ablation study on S2-STDP+NCG. \textit{M} is the number of neurons per class, \textit{CR} is competition regulation with 1 or 2 thresholds, \textit{L} is neuron labeling, and \textit{Drop} is dropout.}
    \begin{subtable}{.5\linewidth}
        \centering
        \footnotesize
        \caption{Fashion-MNIST}
        \begin{tabular}{lcc}
            \toprule
            \multirow{2}{*}{Method} & \multicolumn{2}{c}{Accuracy (Mean\(\pm\)Std \%)}                                        \\
                                    & STDP-CSNN                                        & SoftHebb-CNN                         \\
            \midrule
            \textit{M-1}            & \(85.89 \pm 0.27\)                               & \(90.61 \pm 0.19\)                   \\
            \textit{M-5}            & \(86.74 \pm 0.25\)                               & \(91.25 \pm 0.20\)                   \\
            \textit{M-5+CR-1}       & \(86.77 \pm 0.22\)                               & \(85.16 \pm 5.34\)                   \\
            \textit{M-5+CR-2}       & \(87.76 \pm 0.16\)                               & \(91.33 \pm 0.22\)                   \\
            \textit{M-5+CR-1+L}     & \(87.14 \pm 0.41\)                               & \(89.24 \pm 0.89\)                   \\
            \textit{M-5+CR-2+L}     & \textbf{88.72} \(\pm\) \textbf{0.23}             & \textbf{91.86} \(\pm\) \textbf{0.14} \\
            \midrule
            \midrule
            \textit{M-5+Drop+L}     & \(87.33 \pm 0.20\)                               & \(91.34 \pm 0.08\)                   \\
            \bottomrule
        \end{tabular}
    \end{subtable}%
    \begin{subtable}{.5\linewidth}
        \centering
        \footnotesize
        \caption{CIFAR-10}
        \begin{tabular}{lcc}
            \toprule
            \multirow{2}{*}{Method} & \multicolumn{2}{c}{Accuracy (Mean\(\pm\)Std \%)}                                        \\
                                    & STDP-CSNN                                        & SoftHebb-CNN                         \\
            \midrule
            \textit{M-1}            & \(61.08 \pm 0.17\)                               & \(76.90 \pm 0.27\)                   \\
            \textit{M-5}            & \(62.61 \pm 0.27\)                               & \(78.29 \pm 0.27\)                   \\
            \textit{M-5+CR-1}       & \(64.73 \pm 0.41\)                               & \(77.35 \pm 0.22\)                   \\
            \textit{M-5+CR-2}       & \(65.51 \pm 0.26\)                               & \(78.78 \pm 0.15\)                   \\
            \textit{M-5+CR-1+L}     & \(65.46 \pm 0.40\)                               & \(78.67 \pm 0.19\)                   \\
            \textit{M-5+CR-2+L}     & \textbf{66.41} \(\pm\) \textbf{0.17}             & \textbf{79.55} \(\pm\) \textbf{0.23} \\
            \midrule
            \midrule
            \textit{M-5+Drop+L}     & \(63.15 \pm 0.11\)                               & \(77.98 \pm 0.21\)                   \\
            \bottomrule
        \end{tabular}
    \end{subtable}
    \label{tab:ablation-s2stdp}
\end{table}
We conduct, in Table~\ref{tab:ablation-s2stdp}, an ablation study on S2-STDP+NCG to evaluate each component of our methods.
\textit{M-1} and \textit{M-5} represent S2-STDP+NCG with \(M=1\) (one neuron per class, which is similar to S2-STDP) and \(M=5\), without competition regulation and neuron labeling.
\textit{CR-1} denotes our competition regulation mechanism with a single threshold per neuron (i.e. \(\theta' = \theta\) in Equation~\ref{eq:cr}), as commonly used in WTA-based SNNs~\citep{leeTrainingDeepSpiking2016,falezImprovingSpikingNeural2019}.
Thresholds are not clipped nor reset between epochs, and the learned values are used for inference.
\textit{CR-2} denotes our competition regulation with two-compartment thresholds.
\textit{L} is neuron labeling.
\textit{Drop} is dropout on the output neurons, an alternative competition regulation mechanism employed with R-STDP~\citep{mozafariFirstspikebasedVisualCategorization2018}.
For each method, we optimized hyperparameters with gridsearch (see Section~\ref{sec:setup-protocol}).
Results on both Fashion-MNIST and CIFAR-10 show that each component of our methods (cf. \textit{M-5}, \textit{CR-2}, \textit{L}) brings an individual and significant accuracy gain.
Competition regulation tends to be crucial for benefiting from improved class separation, especially with STDP-CSNN.
The accuracy gain gets lower with SoftHebb-CNN as the extracted features exhibit higher class separability.
Neuron labeling enhances the performance of our models through neuron specialization.
Our competition regulation mechanism based on two-compartment thresholds (cf. \textit{CR-2}) outperforms the existing threshold adaptation with one threshold (cf. \textit{CR-1}), as well as dropout (cf. \textit{Drop}).
In a first-spike-based decision-making context, we find that learning thresholds (cf. \textit{CR-1}) is not mandatory for successfully learning various patterns.
Instead, it is more important to ensure fair decision-making with fixed thresholds and use threshold adaptation as a competition regulation mechanism only.
In Supplementary Material, we provide an ablation study on SSTDP+NCG with similar results (Section~3.4), as well as additional studies on the impact of neuron labeling (Section~3.1) and hyperparameters (Section~3.3).

\subsection{Impact of Competition Regulation} \label{sec:cr-analysis}
\begin{figure}[ht]
    \begin{center}
        \begin{subfigure}[b]{0.5\linewidth}
            \centering
            \includegraphics[width=0.99\linewidth]{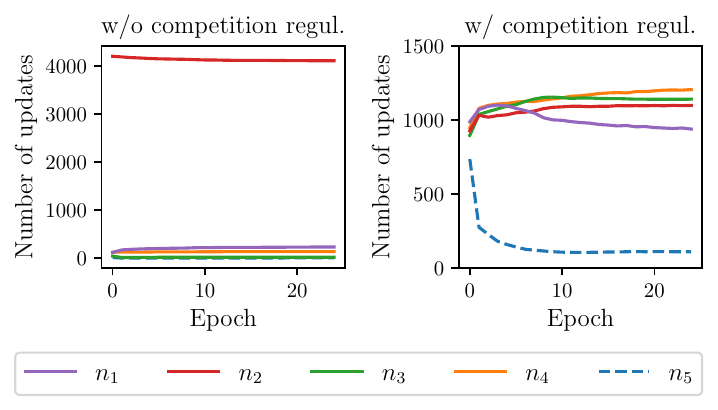}
            \caption{Number of target updates}
        \end{subfigure}%
        \begin{subfigure}[b]{0.5\linewidth}
            \centering
            \includegraphics[width=0.99\linewidth]{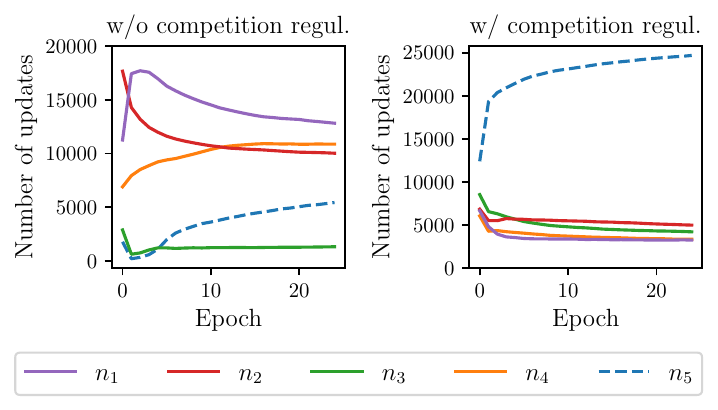}
            \caption{Number of non-target updates}
        \end{subfigure}
    \end{center}
    \caption{Number of weight updates per epoch received by the neurons of class \(0\) trained with S2-STDP+NCG, with and without competition regulation, on CIFAR-10. \(n_1\) to \(n_4\) are labeled as target neurons and \(n_5\) is labeled as non-target. The features are extracted with STDP-CSNN.}
    \label{fig:cifar10-n-updates-s2stdp}
\end{figure}
In this section, we show that competition regulation is crucial for ensuring balanced competition and improved class separation.
Figure~\ref{fig:cifar10-n-updates-s2stdp} illustrates the number of updates per epoch received by the neurons of class \(0\) trained with S2-STDP+NCG, with and without competition regulation, on CIFAR-10.
Target (resp. non-target) updates are triggered by samples of the class (resp. another class).
Without competition regulation, no competition takes place between target neurons.
Target neuron \(n_2\) receives the majority of the target updates, while the other target neurons \(n_1\), \(n_3\), \(n_4\) assume the role of non-target neurons (i.e. receive mainly non-target updates), as they are inhibited by \(n_2\) on samples of the class.
With competition regulation, the target neurons (\(n_1\) to \(n_4\)) effectively specialize toward samples of their class, while the non-target neuron (\(n_5\))  specializes toward samples of other classes.
Regarding target neurons, we observe a balanced competition in their target updates, illustrating the effectiveness of our competition regulation mechanism.
In Supplementary Material (Section~3.5), we show similar results for other classes and datasets, as well as for SSTDP+NCG.

\begin{wrapfigure}{r}{0.46\linewidth}
    \centering%
    \includegraphics[width=1\linewidth]{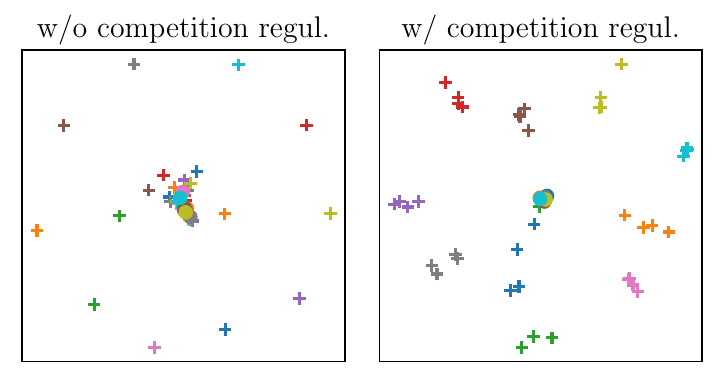}%
    \caption{t-SNE plots of the weights learned with S2-STDP+NCG on CIFAR-10, with and without competition regulation. Crosses and circles respectively represent the weights of target and non-target neurons, and colors indicate classes. The features are extracted with STDP-CSNN.}%
    \label{fig:cifar10-tsne-s2stdp}%
\end{wrapfigure}
In another experiment, we analyze the weights trained using S2-STDP+NCG, with and without competition regulation.
Figure~\ref{fig:cifar10-tsne-s2stdp} shows t-SNE~\citep{maatenVisualizingDataUsing2008} visualizations of the learned weights on CIFAR-10.
Without competition regulation, there is a single cluster at the center, comprising the weights of the neurons that receive few target updates during training.
With competition regulation, the weights of the target neurons tend to form, for each class, distinct clusters.
The spread of their clusters suggests that they have learned various class-specific patterns.
The weights of the non-target neurons form a single cluster at the center since their weights are very similar.
This similarity arises because non-target neurons, regardless of their class, are trained to fire at the same desired timestamp.
In Supplementary Material (Section~3.5), we further show that competition regulation increases the intra-class distinctiveness among weights, which improves class separation.

\section{Discussion} \label{sec:discussion}
The NCG architecture implements effective intra-class WTA in a spiking classification layer employing first-spike coding and supervised STDP training.
Our competition regulation mechanism based on two-compartment thresholds ensures both balanced competition and fair decision-making.
We showed that this mechanism improves class separation and achieves better performance than existing regulation methods.
As a result, NCGs significantly increased the accuracy of SOTA supervised STDP rules.
This work highlights that more effective supervised competitive learning can be achieved without inter-class WTA.
Also, the success in learning various patterns per class via threshold adaptation does not depend on learning various thresholds.

In this work, supervised STDP rules are employed in the classification layer to ensure consistency with the training of the feature extraction network.
However, our contributions focus on the architecture of the classification layer, which is independent of the learning rule used to train it.
Thus, NCGs may theoretically be used with any other rule designed for training SNNs with one spike per neuron.
We performed preliminary experiments with S4NN~\citep{kheradpishehTemporalBackpropagationSpiking2020}, a gradient-based rule, and observed that the addition of NCGs led to accuracy improvements consistent with our previous results (see Section 3.6 of Supplementary Material).
Yet, further research is required to validate the effectiveness of NCGs with gradient-based rules.

While NCGs successfully improve the performance of a classification layer, they also come with several limitations.
First, they increase the costs in terms of parameters, computation, and hardware. The computational overhead of NCGs scales linearly with the number of neurons.
In hardware design, they introduce another overhead due to the additional connections, both to the previous layer and within the layer.
Second, increasing the number of neurons strengthens specialization on the training set, especially when faced with a higher number of input features (cf. SoftHebb-CNN).
This behavior limits the generalization capabilities of our models and requires additional research to fully exploit their potential in these scenarios.
Third, the NCG architecture applies only to the output layer of a network.
Nevertheless, this work is the first to introduce WTA and competition regulation mechanisms specifically designed for classification.
It establishes the relevance of such mechanisms in this context, laying the foundations for future research on WTA-based supervised competitive learning in multi-layer networks.
\vfill

\section*{Acknowledgements}
This work is funded by Chaire Luxant-ANVI (Métropole de Lille) and supported by IRCICA (CNRS UAR 3380).
Experiments presented in this paper were carried out using the Grid'5000 testbed, supported by a scientific interest group hosted by Inria and including CNRS, RENATER and several Universities as well as other organizations (see \url{https://www.grid5000.fr}).

\bibliographystyle{unsrtnat}
\bibliography{references.bib}

\begin{thebibliography}{57}
\providecommand{\natexlab}[1]{#1}
\providecommand{\url}[1]{\texttt{#1}}
\expandafter\ifx\csname urlstyle\endcsname\relax
  \providecommand{\doi}[1]{doi: #1}\else
  \providecommand{\doi}{doi: \begingroup \urlstyle{rm}\Url}\fi

\bibitem[Schuman et~al.(2017)Schuman, Potok, Patton, Birdwell, Dean, Rose, and
  Plank]{schumanSurveyNeuromorphicComputing2017}
Catherine~D. Schuman, Thomas~E. Potok, Robert~M. Patton, J.~Douglas Birdwell,
  Mark~E. Dean, Garrett~S. Rose, and James~S. Plank.
\newblock A {{Survey}} of {{Neuromorphic Computing}} and {{Neural Networks}} in
  {{Hardware}}.
\newblock \emph{ArXiv}, arXiv:1705.06963 [cs.NE], 2017.

\bibitem[Ponulak and Kasinski(2011)]{ponulakIntroductionSpikingNeural2011}
Filip Ponulak and Andrzej Kasinski.
\newblock Introduction to {{Spiking Neural Networks}}: {{Information
  Processing}}, {{Learning}} and {{Applications}}.
\newblock \emph{Acta Neurobiologiae Experimentalis}, 71:\penalty0 409--433,
  2011.

\bibitem[Zou et~al.(2021)Zou, Xu, Chen, Yan, and
  Han]{zouBreakingNeumannBottleneck2021}
Xingqi Zou, Sheng Xu, Xiaoming Chen, Liang Yan, and Yinhe Han.
\newblock Breaking the {{Von Neumann Bottleneck}}: {{Architecture-Level
  Processing-in-Memory Technology}}.
\newblock \emph{Science China Information Sciences}, 64, 2021.

\bibitem[Zenke and Neftci(2021)]{zenkeBrainInspiredLearningNeuromorphic2021}
Friedemann Zenke and Emre Neftci.
\newblock Brain-{{Inspired Learning}} on {{Neuromorphic Substrates}}.
\newblock \emph{Proceedings of the IEEE}, 109:\penalty0 935--950, 2021.

\bibitem[Eshraghian et~al.(2021)Eshraghian, Ward, Neftci, Wang, Lenz, Dwivedi,
  Bennamoun, Jeong, and Lu]{eshraghianTrainingSpikingNeural2021}
Jason~K. Eshraghian, Max Ward, Emre Neftci, Xinxin Wang, Gregor Lenz, Girish
  Dwivedi, Mohammed Bennamoun, Doo~Seok Jeong, and Wei~D. Lu.
\newblock Training {{Spiking Neural Networks Using Lessons From Deep
  Learning}}.
\newblock \emph{ArXiv}, arXiv:2109.12894 [cs.NE], 2021.

\bibitem[Dampfhoffer et~al.(2023)Dampfhoffer, Mesquida, Valentian, and
  Anghel]{dampfhofferBackpropagationBasedLearningTechniques2023}
Manon Dampfhoffer, Thomas Mesquida, Alexandre Valentian, and Lorena Anghel.
\newblock Backpropagation-{{Based Learning Techniques}} for {{Deep Spiking
  Neural Networks}}: {{A Survey}}.
\newblock \emph{Transactions on Neural Networks and Learning Systems}, 2023.

\bibitem[Lillicrap et~al.(2020)Lillicrap, Santoro, Marris, Akerman, and
  Hinton]{lillicrapBackpropagationBrain2020}
Timothy~P. Lillicrap, Adam Santoro, Luke Marris, Colin~J. Akerman, and Geoffrey
  Hinton.
\newblock Backpropagation and the {{Brain}}.
\newblock \emph{Nature Reviews Neuroscience}, 21:\penalty0 335--346, 2020.

\bibitem[Bengio et~al.(2013)Bengio, Courville, and
  Vincent]{bengioRepresentationLearningReview2013}
Yoshua Bengio, Aaron Courville, and Pascal Vincent.
\newblock Representation {{Learning}}: {{A Review}} and {{New Perspectives}}.
\newblock \emph{Transactions on Pattern Analysis and Machine Intelligence},
  35:\penalty0 1798--1828, 2013.

\bibitem[Hebb(1949)]{hebbOrganizationBehavior1949}
Donald Hebb.
\newblock \emph{The {{Organization}} of {{Behavior}}}.
\newblock Springer, Berlin, Heidelberg, 1949.

\bibitem[Caporale and Dan(2008)]{caporaleSpikeTimingDependent2008}
Natalia Caporale and Yang Dan.
\newblock Spike {{Timing}}--{{Dependent Plasticity}}: {{A Hebbian Learning
  Rule}}.
\newblock \emph{Annual Review of Neuroscience}, 31:\penalty0 25--46, 2008.

\bibitem[Sa{\"i}ghi et~al.(2015)Sa{\"i}ghi, Mayr, {Serrano-Gotarredona},
  Schmidt, Lecerf, Tomas, Grollier, Boyn, Vincent, Querlioz, La~Barbera,
  Alibart, Vuillaume, Bichler, Gamrat, and
  {Linares-Barranco}]{saighiPlasticityMemristiveDevices2015}
Sylvain Sa{\"i}ghi, Christian~G. Mayr, Teresa {Serrano-Gotarredona}, Heidemarie
  Schmidt, Gwendal Lecerf, Jean Tomas, Julie Grollier, S{\"o}ren Boyn,
  Adrien~F. Vincent, Damien Querlioz, Selina La~Barbera, Fabien Alibart,
  Dominique Vuillaume, Olivier Bichler, Christian Gamrat, and Bernab{\'e}
  {Linares-Barranco}.
\newblock Plasticity in {{Memristive Devices}} for {{Spiking Neural Networks}}.
\newblock \emph{Frontiers in Neuroscience}, 9, 2015.

\bibitem[Khacef et~al.(2023)Khacef, Klein, Cartiglia, Rubino, Indiveri, and
  Chicca]{khacefSpikebasedLocalSynaptic2023}
Lyes Khacef, Philipp Klein, Matteo Cartiglia, Arianna Rubino, Giacomo Indiveri,
  and Elisabetta Chicca.
\newblock Spike-{{Based Local Synaptic Plasticity}}: {{A Survey}} of
  {{Computational Models}} and {{Neuromorphic Circuits}}.
\newblock \emph{Neuromorphic Computing and Engineering}, 3, 2023.

\bibitem[Falez(2019)]{falezImprovingSpikingNeural2019}
Pierre Falez.
\newblock \emph{Improving {{Spiking Neural Networks Trained}} with {{Spike
  Timing Dependent Plasticity}} for {{Image Recognition}}}.
\newblock PhD thesis, Universit{\'e} de Lille, 2019.

\bibitem[Guo et~al.(2021)Guo, Fouda, Eltawil, and
  Salama]{guoNeuralCodingSpiking2021}
Wenzhe Guo, Mohammed~E. Fouda, Ahmed~M. Eltawil, and Khaled~Nabil Salama.
\newblock Neural {{Coding}} in {{Spiking Neural Networks}}: {{A Comparative
  Study}} for {{Robust Neuromorphic Systems}}.
\newblock \emph{Frontiers in Neuroscience}, 15, 2021.

\bibitem[Rueckauer and Liu(2018)]{rueckauerConversionAnalogSpiking2018}
Bodo Rueckauer and Shih-Chii Liu.
\newblock Conversion of {{Analog}} to {{Spiking Neural Networks Using Sparse
  Temporal Coding}}.
\newblock In \emph{International {{Symposium}} on {{Circuits}} and
  {{Systems}}}, 2018.

\bibitem[Park et~al.(2020)Park, Kim, Na, and Yoon]{parkT2FSNNDeepSpiking2020}
Seongsik Park, Seijoon Kim, Byunggook Na, and Sungroh Yoon.
\newblock {{T2FSNN}}: {{Deep Spiking Neural Networks}} with
  {{Time-to-First-Spike Coding}}.
\newblock In \emph{Design {{Automation Conference}}}, 2020.

\bibitem[Rullen and Thorpe(2001)]{rullenRateCodingTemporal2001}
Rufin~Van Rullen and Simon~J. Thorpe.
\newblock Rate {{Coding Versus Temporal Order Coding}}: {{What}} the {{Retinal
  Ganglion Cells Tell}} the {{Visual Cortex}}.
\newblock \emph{Neural Computation}, 13:\penalty0 1255--1283, 2001.

\bibitem[Auge et~al.(2021)Auge, Hille, Mueller, and
  Knoll]{augeSurveyEncodingTechniques2021}
Daniel Auge, Julian Hille, Etienne Mueller, and Alois Knoll.
\newblock A {{Survey}} of {{Encoding Techniques}} for {{Signal Processing}} in
  {{Spiking Neural Networks}}.
\newblock \emph{Neural Processing Letters}, 53:\penalty0 4693--4710, 2021.

\bibitem[Kheradpisheh et~al.(2018)Kheradpisheh, Ganjtabesh, Thorpe, and
  Masquelier]{kheradpishehSTDPbasedSpikingDeep2018}
Saeed~Reza Kheradpisheh, Mohammad Ganjtabesh, Simon~J. Thorpe, and Timoth{\'e}e
  Masquelier.
\newblock {{STDP-Based Spiking Deep Convolutional Neural Networks}} for
  {{Object Recognition}}.
\newblock \emph{Neural Networks}, 99:\penalty0 56--67, 2018.

\bibitem[Falez et~al.(2019)Falez, Tirilly, Marius~Bilasco, Devienne, and
  Boulet]{falezMultiLayeredSpikingNeural2019}
Pierre Falez, Pierre Tirilly, Ioan Marius~Bilasco, Philippe Devienne, and
  Pierre Boulet.
\newblock Multi-{{Layered Spiking Neural Network}} with {{Target Timestamp
  Threshold Adaptation}} and {{STDP}}.
\newblock In \emph{International {{Joint Conference}} on {{Neural Networks}}},
  2019.

\bibitem[{El-Assal} et~al.(2022){El-Assal}, Tirilly, and
  Bilasco]{el-assal2D3DConvolutional2022}
Mireille {El-Assal}, Pierre Tirilly, and Ioan~Marius Bilasco.
\newblock {{2D Versus 3D Convolutional Spiking Neural Networks Trained}} with
  {{Unsupervised STDP}} for {{Human Action Recognition}}.
\newblock In \emph{International {{Joint Conference}} on {{Neural Networks}}},
  2022.

\bibitem[Mozafari et~al.(2018)Mozafari, Kheradpisheh, Masquelier,
  {Nowzari-Dalini}, and
  Ganjtabesh]{mozafariFirstspikebasedVisualCategorization2018}
Milad Mozafari, Saeed~Reza Kheradpisheh, Timothee Masquelier, Abbas
  {Nowzari-Dalini}, and Mohammad Ganjtabesh.
\newblock First-{{Spike-Based Visual Categorization Using Reward-Modulated
  STDP}}.
\newblock \emph{Transactions on Neural Networks and Learning Systems},
  29:\penalty0 6178--6190, 2018.

\bibitem[Zhao et~al.(2020)Zhao, Zeng, Zhang, Shi, and
  Zhao]{zhaoGLSNNMultiLayerSpiking2020}
Dongcheng Zhao, Yi~Zeng, Tielin Zhang, Mengting Shi, and Feifei Zhao.
\newblock {{GLSNN}}: {{A Multi-Layer Spiking Neural Network Based}} on {{Global
  Feedback Alignment}} and {{Local STDP Plasticity}}.
\newblock \emph{Frontiers in Computational Neuroscience}, 14, 2020.

\bibitem[Liu et~al.(2021)Liu, Zhao, Chen, Wang, Yang, and
  Jiang]{liuSSTDPSupervisedSpike2021}
Fangxin Liu, Wenbo Zhao, Yongbiao Chen, Zongwu Wang, Tao Yang, and Li~Jiang.
\newblock {{SSTDP}}: {{Supervised Spike Timing Dependent Plasticity}} for
  {{Efficient Spiking Neural Network Training}}.
\newblock \emph{Frontiers in Neuroscience}, 15, 2021.

\bibitem[Shrestha et~al.(2017)Shrestha, Ahmed, Wang, and
  Qiu]{shresthaStableSpiketimingDependent2017}
Amar Shrestha, Khadeer Ahmed, Yanzhi Wang, and Qinru Qiu.
\newblock Stable {{Spike-Timing Dependent Plasticity Rule}} for {{Multilayer
  Unsupervised}} and {{Supervised Learning}}.
\newblock In \emph{International {{Joint Conference}} on {{Neural Networks}}},
  pages 1999--2006, 2017.

\bibitem[Mozafari et~al.(2019)Mozafari, Ganjtabesh, {Nowzari-Dalini}, Thorpe,
  and Masquelier]{mozafariBioinspiredDigitRecognition2019}
Milad Mozafari, Mohammad Ganjtabesh, Abbas {Nowzari-Dalini}, Simon~J. Thorpe,
  and Timoth{\'e}e Masquelier.
\newblock Bio-{{Inspired Digit Recognition Using Reward-Modulated
  Spike-Timing-Dependent Plasticity}} in {{Deep Convolutional Networks}}.
\newblock \emph{Pattern Recognition}, 94, 2019.

\bibitem[Thiele et~al.(2018)Thiele, Bichler, and
  Dupret]{thieleEventBasedTimescaleInvariant2018}
Johannes~C. Thiele, Olivier Bichler, and Antoine Dupret.
\newblock Event-{{Based}}, {{Timescale Invariant Unsupervised Online Deep
  Learning}} with {{STDP}}.
\newblock \emph{Frontiers in Computational Neuroscience}, 12, 2018.

\bibitem[Lee et~al.(2019)Lee, Srinivasan, Panda, and
  Roy]{leeDeepSpikingConvolutional2019}
Chankyu Lee, Gopalakrishnan Srinivasan, Priyadarshini Panda, and Kaushik Roy.
\newblock Deep {{Spiking Convolutional Neural Network Trained}} with
  {{Unsupervised Spike-Timing-Dependent Plasticity}}.
\newblock \emph{Transactions on Cognitive and Developmental Systems},
  11:\penalty0 384--394, 2019.

\bibitem[Goupy et~al.(2024)Goupy, Tirilly, and
  Bilasco]{goupyPairedCompetingNeurons2024}
Gaspard Goupy, Pierre Tirilly, and Ioan~Marius Bilasco.
\newblock Paired {{Competing Neurons Improving STDP Supervised Local Learning}}
  in {{Spiking Neural Networks}}.
\newblock \emph{Frontiers in Neuroscience}, 18, 2024.

\bibitem[Diehl and Cook(2015)]{diehlUnsupervisedLearningDigit2015}
Peter Diehl and Matthew Cook.
\newblock Unsupervised {{Learning}} of {{Digit Recognition Using
  Spike-Timing-Dependent Plasticity}}.
\newblock \emph{Frontiers in Computational Neuroscience}, 9, 2015.

\bibitem[Ferr{\'e} et~al.(2018)Ferr{\'e}, Mamalet, and
  Thorpe]{ferreUnsupervisedFeatureLearning2018}
Paul Ferr{\'e}, Franck Mamalet, and Simon~J. Thorpe.
\newblock Unsupervised {{Feature Learning}} with {{Winner-Takes-All Based
  STDP}}.
\newblock \emph{Frontiers in Computational Neuroscience}, 12, 2018.

\bibitem[Lee et~al.(2016)Lee, Delbruck, and
  Pfeiffer]{leeTrainingDeepSpiking2016}
Jun~Haeng Lee, Tobi Delbruck, and Michael Pfeiffer.
\newblock Training {{Deep Spiking Neural Networks Using Backpropagation}}.
\newblock \emph{Frontiers in Neuroscience}, 10, 2016.

\bibitem[Qu et~al.(2020)Qu, Zhao, Wang, and
  Wang]{quEfficientHardwareFriendlyMethods2020}
Lianhua Qu, Zhenyu Zhao, Lei Wang, and Yong Wang.
\newblock Efficient and {{Hardware-Friendly Methods}} to {{Implement
  Competitive Learning}} for {{Spiking Neural Networks}}.
\newblock \emph{Neural Computing and Applications}, 32, 2020.

\bibitem[Hao et~al.(2020)Hao, Huang, Dong, and
  Xu]{haoBiologicallyPlausibleSupervised2020}
Yunzhe Hao, Xuhui Huang, Meng Dong, and Bo~Xu.
\newblock A {{Biologically Plausible Supervised Learning Method}} for {{Spiking
  Neural Networks Using}} the {{Symmetric STDP Rule}}.
\newblock \emph{Neural Networks}, 121:\penalty0 387--395, 2020.

\bibitem[Bethi et~al.(2022)Bethi, Xu, Cohen, Van~Schaik, and
  Afshar]{bethiOptimizedDeepSpiking2022}
Yeshwanth Bethi, Ying Xu, Gregory Cohen, Andr{\'e} Van~Schaik, and Saeed
  Afshar.
\newblock An {{Optimized Deep Spiking Neural Network Architecture Without
  Gradients}}.
\newblock \emph{IEEE Access}, 10:\penalty0 97912--97929, 2022.

\bibitem[Fr{\'e}maux and
  Gerstner(2015)]{fremauxNeuromodulatedSpikeTimingDependentPlasticity2015}
Nicolas Fr{\'e}maux and Wulfram Gerstner.
\newblock Neuromodulated {{Spike-Timing-Dependent Plasticity}}, and {{Theory}}
  of {{Three-Factor Learning Rules}}.
\newblock \emph{Frontiers in Neural Circuits}, 9, 2015.

\bibitem[Ponulak and Kasi{\'n}ski(2010)]{ponulakSupervisedLearningSpiking2010}
Filip Ponulak and Andrzej Kasi{\'n}ski.
\newblock Supervised {{Learning}} in {{Spiking Neural Networks}} with
  {{ReSuMe}}: {{Sequence Learning}}, {{Classification}}, and {{Spike
  Shifting}}.
\newblock \emph{Neural Computation}, 22:\penalty0 467--510, 2010.

\bibitem[Tavanaei and
  Maida(2019)]{tavanaeiBPSTDPApproximatingBackpropagation2019}
Amirhossein Tavanaei and Anthony Maida.
\newblock {{BP-STDP}}: {{Approximating Backpropagation Using Spike Timing
  Dependent Plasticity}}.
\newblock \emph{Neurocomputing}, 330:\penalty0 39--47, 2019.

\bibitem[Shrestha et~al.(2019)Shrestha, Fang, Wu, and
  Qiu]{shresthaApproximatingBackPropagationBiologically2019}
Amar Shrestha, Haowen Fang, Qing Wu, and Qinru Qiu.
\newblock Approximating {{Back-Propagation}} for a {{Biologically Plausible
  Local Learning Rule}} in {{Spiking Neural Networks}}.
\newblock In \emph{International {{Conference}} on {{Neuromorphic Systems}}},
  2019.

\bibitem[Saranirad et~al.(2022)Saranirad, Dora, McGinnity, and
  Coyle]{saraniradAssemblybasedSTDPNew2022}
Vahid Saranirad, Shirin Dora, T.~M. McGinnity, and Damien Coyle.
\newblock Assembly-{{Based STDP}}: {{A New Learning Rule}} for {{Spiking Neural
  Networks Inspired}} by {{Biological Assemblies}}.
\newblock In \emph{International {{Joint Conference}} on {{Neural Networks}}},
  2022.

\bibitem[Beyeler et~al.(2013)Beyeler, Dutt, and
  Krichmar]{beyelerCategorizationDecisionmakingNeurobiologically2013}
Michael Beyeler, Nikil~D. Dutt, and Jeffrey~L. Krichmar.
\newblock Categorization and {{Decision-Making}} in a {{Neurobiologically
  Plausible Spiking Network Using}} a {{STDP-Like Learning Rule}}.
\newblock \emph{Neural Networks}, 48:\penalty0 109--124, 2013.

\bibitem[Luo et~al.(2019)Luo, Qu, Zhang, and
  Chen]{luoFirstErrorBasedSupervised2019}
Xiaoling Luo, Hong Qu, Yun Zhang, and Yi~Chen.
\newblock First {{Error-Based Supervised Learning Algorithm}} for {{Spiking
  Neural Networks}}.
\newblock \emph{Frontiers in Neuroscience}, 13, 2019.

\bibitem[Wang et~al.(2021)Wang, Shi, Zhou, Lin, He, Gan, Li, Wang, Liu, Wu, and
  Luo]{wangCompSNNLightweightSpiking2021}
Tengxiao Wang, Cong Shi, Xichuan Zhou, Yingcheng Lin, Junxian He, Ping Gan,
  Ping Li, Ying Wang, Liyuan Liu, Nanjian Wu, and Gang Luo.
\newblock {{CompSNN}}: {{A Lightweight Spiking Neural Network Based}} on
  {{Spatiotemporally Compressive Spike Features}}.
\newblock \emph{Neurocomputing}, 425:\penalty0 96--106, 2021.

\bibitem[Kulkarni and Rajendran(2018)]{kulkarniSpikingNeuralNetworks2018}
Shruti~R. Kulkarni and Bipin Rajendran.
\newblock Spiking {{Neural Networks}} for {{Handwritten Digit
  Recognition}}---{{Supervised Learning}} and {{Network Optimization}}.
\newblock \emph{Neural Networks}, 103:\penalty0 118--127, 2018.

\bibitem[Lobov et~al.(2020)Lobov, Chernyshov, Krilova, Shamshin, and
  Kazantsev]{lobovCompetitiveLearningSpiking2020}
Sergey~A. Lobov, Andrey~V. Chernyshov, Nadia~P. Krilova, Maxim~O. Shamshin, and
  Victor~B. Kazantsev.
\newblock Competitive {{Learning}} in a {{Spiking Neural Network}}: {{Towards}}
  an {{Intelligent Pattern Classifier}}.
\newblock \emph{Sensors}, 20, 2020.

\bibitem[Srivastava et~al.(2014)Srivastava, Hinton, Krizhevsky, Sutskever, and
  Salakhutdinov]{srivastavaDropoutSimpleWay2014}
Nitish Srivastava, Geoffrey Hinton, Alex Krizhevsky, Ilya Sutskever, and Ruslan
  Salakhutdinov.
\newblock Dropout: {{A Simple Way}} to {{Prevent Neural Networks}} from
  {{Overfitting}}.
\newblock \emph{Journal of Machine Learning Research}, 15:\penalty0 1929--1958,
  2014.

\bibitem[Yu et~al.(2022)Yu, Ma, Song, Zhang, Dang, and
  Tan]{yuConstructingAccurateEfficient2022}
Qiang Yu, Chenxiang Ma, Shiming Song, Gaoyan Zhang, Jianwu Dang, and Kay~Chen
  Tan.
\newblock Constructing {{Accurate}} and {{Efficient Deep Spiking Neural
  Networks}} with {{Double-Threshold}} and {{Augmented Schemes}}.
\newblock \emph{Transactions on Neural Networks and Learning Systems},
  33:\penalty0 1714--1726, 2022.

\bibitem[Goupy et~al.(2023)Goupy, {Juneau-Fecteau}, Garg, Balafrej, Alibart,
  Frechette, Drouin, and Beilliard]{goupyUnsupervisedEfficientLearning2023}
Gaspard Goupy, Alexandre {Juneau-Fecteau}, Nikhil Garg, Ismael Balafrej, Fabien
  Alibart, Luc Frechette, Dominique Drouin, and Yann Beilliard.
\newblock Unsupervised and {{Efficient Learning}} in {{Sparsely Activated
  Convolutional Spiking Neural Networks Enabled}} by {{Voltage-Dependent
  Synaptic Plasticity}}.
\newblock \emph{Neuromorphic Computing and Engineering}, 3, 2023.

\bibitem[LeCun et~al.(1998)LeCun, Bottou, Bengio, and
  Haffner]{lecunGradientbasedLearningApplied1998}
Yann LeCun, L{\'e}on Bottou, Yoshua Bengio, and Patrick Haffner.
\newblock Gradient-{{Based Learning Applied}} to {{Document Recognition}}.
\newblock \emph{Proceedings of the IEEE}, 86:\penalty0 2278--2323, 1998.

\bibitem[Xiao et~al.(2017)Xiao, Rasul, and
  Vollgraf]{xiaoFashionMNISTNovelImage2017}
Han Xiao, Kashif Rasul, and Roland Vollgraf.
\newblock Fashion-{{MNIST}}: A {{Novel Image Dataset}} for {{Benchmarking
  Machine Learning Algorithms}}.
\newblock \emph{ArXiv}, arXiv:1708.07747 [cs.LG], 2017.

\bibitem[Krizhevsky(2009)]{krizhevskyLearningMultipleLayers2009}
Alex Krizhevsky.
\newblock Learning {{Multiple Layers}} of {{Features}} from {{Tiny Images}}.
\newblock Technical report, University of Toronto, USA, 2009.

\bibitem[Journ{\'e} et~al.(2023)Journ{\'e}, Rodriguez, Guo, and
  Moraitis]{journeHebbianDeepLearning2023}
Adrien Journ{\'e}, Hector~Garcia Rodriguez, Qinghai Guo, and Timoleon Moraitis.
\newblock Hebbian {{Deep Learning Without Feedback}}.
\newblock \emph{International Conference on Learning Representations}, 2023.

\bibitem[Thorpe et~al.(2001)Thorpe, Delorme, and
  Van~Rullen]{thorpeSpikebasedStrategiesRapid2001}
Simon Thorpe, Arnaud Delorme, and Rufin Van~Rullen.
\newblock Spike-{{Based Strategies}} for {{Rapid Processing}}.
\newblock \emph{Neural Networks}, 14:\penalty0 715--725, 2001.

\bibitem[Mirsadeghi et~al.(2023)Mirsadeghi, Shalchian, Kheradpisheh, and
  Masquelier]{mirsadeghiSpikeTimeDisplacementbased2023}
Maryam Mirsadeghi, Majid Shalchian, Saeed~Reza Kheradpisheh, and Timoth{\'e}e
  Masquelier.
\newblock Spike {{Time Displacement-Based Error Backpropagation}} in
  {{Convolutional Spiking Neural Networks}}.
\newblock \emph{Neural Computing and Applications}, 35:\penalty0 15891--15906,
  2023.

\bibitem[Shrestha et~al.(2021)Shrestha, Fang, Rider, Mei, and
  Qiu]{shresthaInHardwareLearningMultilayer2021}
Amar Shrestha, Haowen Fang, Daniel~Patrick Rider, Zaidao Mei, and Qinru Qiu.
\newblock In-{{Hardware Learning}} of {{Multilayer Spiking Neural Networks}} on
  a {{Neuromorphic Processor}}.
\newblock In \emph{Design {{Automation Conference}}}, pages 367--372, 2021.

\bibitem[van~der Maaten and Hinton(2008)]{maatenVisualizingDataUsing2008}
Laurens van~der Maaten and Geoffrey Hinton.
\newblock Visualizing {{Data Using}} t-{{SNE}}.
\newblock \emph{Journal of Machine Learning Research}, 9:\penalty0 2579--2605,
  2008.

\bibitem[Kheradpisheh and
  Masquelier(2020)]{kheradpishehTemporalBackpropagationSpiking2020}
Saeed~Reza Kheradpisheh and Timoth{\'e}e Masquelier.
\newblock Temporal {{Backpropagation}} for {{Spiking Neural Networks}} with
  {{One Spike}} per {{Neuron}}.
\newblock \emph{International Journal of Neural Systems}, 30, 2020.

\end{thebibliography}




\section*{Supplementary material}

# Supplementary Material

Gaspard Goupy[1], Pierre Tirilly[1], and Ioan Marius Bilasco[1]

[1]Univ. Lille, CNRS, Centrale Lille, UMR 9189 CRIStAL, F-59000 Lille, France

## 1 Overall Algorithm

We present, in Algorithm 1, the pseudo-code for training a spiking classification layer with S2-STDP+NCG. We omitted early stopping for clarity, as well as how we handle intra-class inhibition if several neurons fire at the same time.

---

**Algorithm 1** Training of a spiking classification layer with S2-STDP+NCG.

---
1: **Input:**
2: Number of samples $D \in \mathbb{Z}^+$;
3: Number of input neurons $F \in \mathbb{Z}^+$;
4: Number of output neurons $N \in \mathbb{Z}^+$;
5: Number of neurons per class $M \in \mathbb{Z}^+$;
6: Number of classes $C \in \mathbb{Z}^+$;
7: Firing threshold $V_{\text{th}} \in \mathbb{R}$;
8: Maximum firing time $T_{\max} \in \mathbb{R}^+$;
9: Initial weight normal distribution of mean $\mu \in \mathbb{R}$ and std $\sigma \in \mathbb{R}_0^+$;
10: Minimum and maximum weight value $w_{\min} \in \mathbb{R}$ and $w_{\max} \in \mathbb{R}$;
11: S2-STDP time gap $g \in \mathbb{R}^+$;
12: STDP learning rates $A^+ \in \mathbb{R}_0^+$ and $A^- \in \mathbb{R}_0^-$;
13: STDP annealing $\beta \in \mathbb{R}_0^+$;
14: Threshold learning rate $\eta_{\text{th}} \in \mathbb{R}_0^+$;
15: Threshold annealing $\beta_{\text{th}} \in \mathbb{R}_0^+$;
16: Matrix of spike-encoded extracted features $X \in [0, T_{\max}]^{D \times F}$;
17: Vector of class labels $Y \in \mathbb{Z}^D$.
18: **Output:** Matrix of trained weights $W \in \mathbb{R}^{F \times N}$.
19: Define $C$ NCGs of $M$ neurons
20: For each NCG, label $M - 1$ neurons as *target* and 1 as *non-target*
21: Map each neuron $n_j$ to a class $c_j \in \mathbb{Z}$ based on its NCG ($j$ from 1 to $N$)
22: Initialize matrix of weights $W \sim \mathcal{N}(\mu, \sigma^2)$
23: Initialize vector of test thresholds $\theta \in \mathbb{R}^N, \theta_j := V_{th}$ ($j$ from 1 to $N$)
24: Initialize vector of training thresholds $\theta' \in \mathbb{R}^N, \theta'_j := V_{th}$ ($j$ from 1 to $N$)
25: Compute vector of weight normalization factors $\overline{w} \in \mathbb{R}^N, \overline{w}_j := \sum_i W_{ij}$ ($j$ from 1 to $N$)
26: **for each** epoch **do**
27:     Reset $\theta'_j$ to $\theta_j$ ($j$ from 1 to $N$)     ▷ Reset training thresholds to test thresholds
28:     **for each** $x, y$ **in** $X, Y$ **do**     ▷ For each sample of data $x$ and class $y$

```
29:        // Forward propagation
30:        Convert x ∈ [0, T_max]^F into an ordered vector of input spike times t^I ∈ [0, T_max]^F
31:        Initialize vector of output spike times t^O ∈ [0, T_max]^N, t_j^O := T_max (j from 1 to N)
32:        // T_max = no spike
33:        Initialize vector of membrane potentials V ∈ ℝ^N, V_j := 0 (j from 1 to N)
34:        Initialize vector of active flags A ∈ {0,1}^N, A_j := 1 (j from 1 to N)
35:        for each t_k^I ∈ t^I do                                    ▷ For each ordered input spike time
36:            for each n_j where A_j = 1 do                          ▷ For each active neuron
37:                V_j := V_j + W_kj                                  ▷ Integrate input spike, Eq. 1 in main
38:                if V_j ≥ (θ_j if c_j ≠ y else θ'_j) then            ▷ Threshold depends on sample class
39:                    t_j^O := t_k^I                                  ▷ n_j fires at t_k^I
40:                    A_p := 0, for each n_p in the NCG of n_j        ▷ Intra-class inhibition
41:                end if
42:            end for
43:        end for
44:        // Weights and thresholds update
45:        T̄ := mean {t ∈ t^O | t < T_max}, T̄ ∈ ℝ  ▷ Average firing time of non-inhibited neurons
46:        for each NCG do                                             ▷ One update per NCG
47:            Get the first n_{j*} to fire within the NCG              ▷ Non-inhibited neuron
48:            if n_{j*} is target and c_{j*} = y then                  ▷ Target update
49:                e_{j*} := t_{j*}^O − (T̄ − (C−1)/C · g), e_{j*} ∈ ℝ    ▷ Eq. 5 in main
50:                for each n_p in the NCG where n_p is target do       ▷ Competition regulation
51:                    if p = j* then                                   ▷ Non-inhibited target neuron
52:                        θ'_p := max {θ_p, θ'_p + η_th · (M−2)/(M−1)}  ▷ Eq. 6 in main
53:                    else                                             ▷ Inhibited target neurons
54:                        θ'_p := max {θ_p, θ'_p − η_th · 1/(M−1)}      ▷ Eq. 6 in main
55:                    end if
56:                end for
57:            else                                                     ▷ Non-target update
58:                e_{j*} := t_{j*}^O − (T̄ + (1/C) g), e_{j*} ∈ ℝ        ▷ Eq. 5 in main
59:            end if
60:            for k ∈ [1, F] do                         ▷ Update weights of n_{j*} with error-modulated STDP
61:                if t_k^I ≤ t_{j*}^O then                              ▷ Long-term potentiation
62:                    W_{kj*} := W_{kj*} + e_{j*} · A^+                 ▷ Eq. 3 in main
63:                else                                                  ▷ Long-term depression
64:                    W_{kj*} := W_{kj*} + e_{j*} · A^−                 ▷ Eq. 3 in main
65:                end if
66:            end for
67:            W_{kj*} := W_{kj*} · w̄_{j*} / Σ_x |W_{xj*}|  (k from 1 to F)  ▷ Weight normalization
68:            Clip weights of n_{j*} in [w_min, w_max]
69:        end for
70:    end for
71:    A^+ := A^+ · β, A^− := A^− · β                                   ▷ STDP learning rate annealing
72:    η_th := η_th · β_th                                              ▷ Threshold learning rate annealing
73: end for
```

## 2 Experimental Details

### 2.1 Classification Pipeline

We present, in Figure 1, the classification pipeline used in our experiments. Our classification system consists of an unsupervised feature extractor followed by a spiking classification layer.

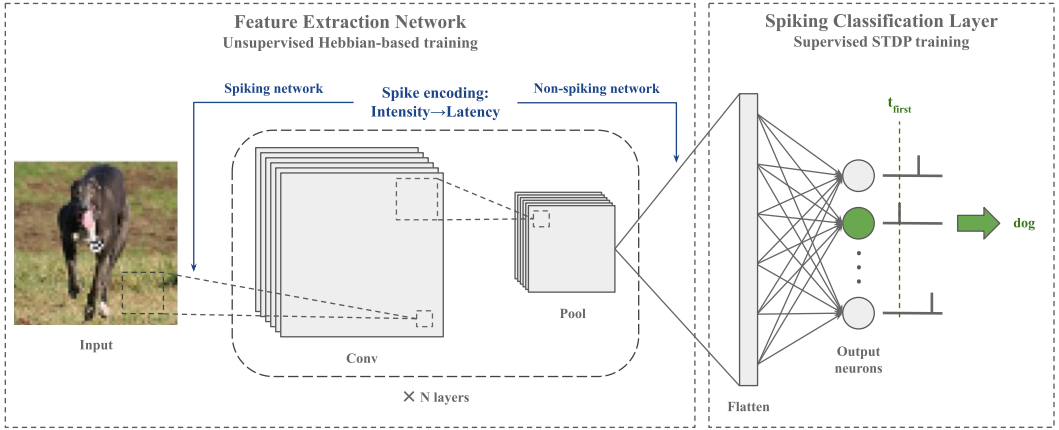

Figure 1: Pipeline of our classification system. Training is performed in a layer-wise fashion. First, the convolutional layers of the feature extractor are trained on the input images using an unsupervised Hebbian-based learning rule. Then, a fully-connected spiking classification layer is trained on the extracted features using a supervised STDP rule. Spike encoding occurs before feature extraction for spike-based feature extractors or afterward for non-spike-based extractors.

### 2.2 Unsupervised Feature Extractors

**STDP-CSNN** comprises a spiking convolutional layer trained with STDP and a spiking max-pooling layer. First, images are preprocessed with on-center/off-center coding [1] for MNIST and Fashion-MNIST, and with hardware-friendly whitening [2] for CIFAR-10/100. Then, preprocessed images are encoded into spikes with latency coding [3] (a form of first-spike coding) and transmitted to STDP-CSNN for unsupervised training. Latency coding converts each normalized pixel $x$ into a single spike timestamp as follows: $T(x) = 1 - x$. Each output feature consists of a floating-point spike timestamp in $[0, 1]$, which is inherently compatible with the classification layer. We use the hyperparameters reported in [4]. The base code is available at https://gitlab.univ-lille.fr/bioinsp/falez-csnn-simulator, under the CeCILL-B license.

**SoftHebb-CNN** comprises three convolutional layers trained with SoftHebb. Each convolutional layer includes a succession of batch normalization, convolution, pooling (max-pooling for the first two layers, average-pooling for the last one), and a Triangle [5] activation function. Images undergo no preprocessing. The extracted output features of each sample are rescaled in $[0, 1]$ and encoded into spike timestamps with latency coding (see supra). For all the datasets, we use the hyperparameters employed for the CIFAR-10 task in the original paper [6]. The base code is available at https://github.com/NeuromorphicComputing/SoftHebb. No license information is available.

### 2.3 Spiking Classification Layers

The maximum firing time in the layer is $T_{\max} = 1$, which corresponds to the maximum possible firing timestamp from input neurons. We initialize weights with a normal distribution of mean $\mu$ and standard deviation $\sigma$, and we clip them in $[w_{\min}, w_{\max}]$ after each update. For SSTDP and S2-STDP models, we use values from [4]: $\mu = 0.5$ for SSTDP and $\mu = 0.3$ for S2-STDP, $\sigma = 0.01$, $w_{\min} = 0$, and $w_{\max} = 1$. For R-STDP models, we use values from [7]: $\mu = 0.8$, $\sigma = 0.01$, $w_{\min} = 0.2$, and $w_{\max} = 0.8$. Weight normalization is employed with S2-STDP models [4] to keep a similar weight average (of $\mu = 0.3$) across neurons during learning. We decrease the learning rates after each epoch by a factor $\beta = 0.98$. All the other hyperparameters (including firing threshold,

learning rates, and method-specific hyperparameters) are optimized with gridsearch. To ensure a fair comparison between methods, we performed extensive gridsearch optimization (from 600 to 1,440 runs, depending on the number of hyperparameters) for each model (except on CIFAR-100 where we used hyperparameters optimized on CIFAR-10). Optimized values and gridsearch ranges are located in the config/ folder of our source code.

### 2.4 Computing Resources

We conducted our experiments on private servers as well as the Grid'5000 testbed [8], providing clusters of servers. Since our models rely on floating-point values to represent firing timestamps, we cannot benefit from GPU parallelization of operations. Therefore, we exclusively used CPUs, primarily Intel Xeon W-2245, Intel Xeon Gold 5218, Intel Xeon Gold 6126, Intel Xeon Gold 6130, Intel Xeon Platinum 8358, AMD EPYC 7343. The servers ranged from 16 to 128 CPU cores and 64 to 768 GiB of RAM. The entire experimentation process spanned over five months and resulted in over 70,000 runs (i.e. training of a single model), including gridsearch runs, K-fold runs, and unreported experiments. The average duration of a single run varies depending on several factors, such as the number of input features and the dataset. On CIFAR-10, training a classification layer with S2-STDP+NCG (total of 50 output neurons) using pre-extracted features requires approximately 1 hour with STDP-CSNN (for $\sim 1.4$ GiB of RAM) and 4 hours with SoftHebb-CNN (for $\sim 8.9$ GiB of RAM). On CIFAR-100, the same training (total of 500 neurons, as ten times more classes) requires approximately 8 hours with STDP-CSNN (for $\sim 1.3$ GiB of RAM) and 48 hours with SoftHebb-CNN (for $\sim 8.7$ GiB of RAM). Note that the amount of RAM needed is mainly influenced by the number of input features. Also, the computational overhead of NCGs scales linearly with both the number of neurons and the number of input features.

### 2.5 Datasets

MNIST is available at http://yann.lecun.com/exdb/mnist/ under the CC BY-SA 3.0 license. Fashion-MNIST is available at https://github.com/zalandoresearch/fashion-mnist under the MIT license. CIFAR-10 and CIFAR-100 are available at https://www.cs.toronto.edu/~kriz/cifar.html under the MIT license.

## 3 Additional Experiments

### 3.1 Impact of the Non-Target Neuron

Table 1: Accuracy of S2-STDP+NCG with various numbers of target and non-target neurons.

| Dataset | Non-target neurons | Target neurons | Accuracy (Mean±Std %) | |
|---|---|---|---|---|
| | | | STDP-CSNN | SoftHebb-CNN |
| MNIST | 1 | 1 | $98.62 \pm 0.07$ | $\mathbf{99.18 \pm 0.05}$ |
| | 0 | 4 | $98.79 \pm 0.07$ | $99.05 \pm 0.07$ |
| | 0 | 5 | $98.82 \pm 0.05$ | $99.08 \pm 0.08$ |
| | 1 | 4 | $\mathbf{98.92 \pm 0.07}$ | $99.17 \pm 0.07$ |
| Fashion-MNIST | 1 | 1 | $87.45 \pm 0.16$ | $91.33 \pm 0.21$ |
| | 0 | 4 | $87.73 \pm 0.24$ | $91.34 \pm 0.17$ |
| | 0 | 5 | $87.76 \pm 0.16$ | $91.33 \pm 0.22$ |
| | 1 | 4 | $\mathbf{88.72 \pm 0.23}$ | $\mathbf{91.86 \pm 0.14}$ |
| CIFAR-10 | 1 | 1 | $62.94 \pm 0.17$ | $78.33 \pm 0.18$ |
| | 0 | 4 | $64.85 \pm 0.16$ | $78.54 \pm 0.26$ |
| | 0 | 5 | $65.51 \pm 0.26$ | $78.78 \pm 0.15$ |
| | 1 | 4 | $\mathbf{66.41 \pm 0.17}$ | $\mathbf{79.55 \pm 0.23}$ |

In [4], the accuracy of a classification layer trained with S2-STDP is improved by using two neurons per class with intra-class WTA competition. The authors show that this accuracy gain is attributed

18

to implicit neuron specialization toward target or non-target samples. In S2-STDP+NCG models, we take advantage of this mechanism by explicitly labeling one neuron of each NCG as non-target, whereas the others are labeled as target. To evaluate the impact of such labeling, we compare, in Table 1, the accuracy of S2-STDP+NCG with various numbers of target and non-target neurons. When using one target and one non-target neuron (similarly to [4] but with explicit labeling), the classification layer cannot learn various patterns per class but the accuracy is still improved through neuron specialization. We tend to obtain similar or better accuracies with multiple target neurons and no non-target neurons by learning various patterns per class. When combining multiple target neurons and a non-target neuron, we reach optimal performance. Preliminary experiments with multiple non-target neurons and competition regulation did not yield improvements in accuracy. We believe that a single non-target neuron is enough to benefit from the specialization effect.

### 3.2 Impact of the Number of Neurons

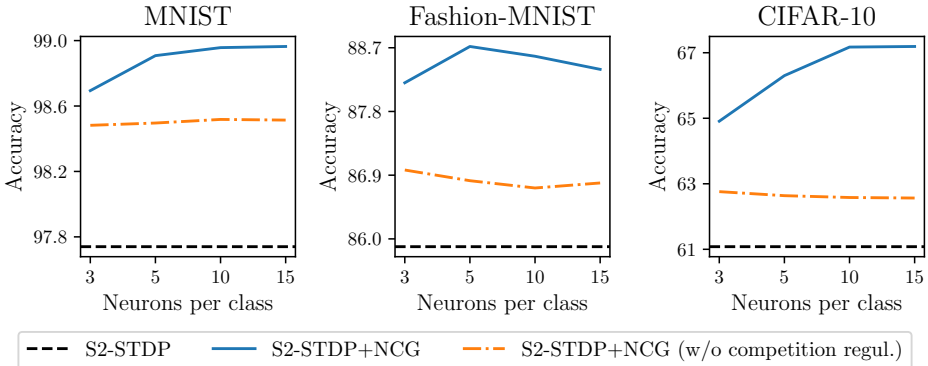

Figure 2: Accuracy against the number of neurons per class in the classification layer trained with S2-STDP+NCG. The features are extracted with STDP-CSNN.

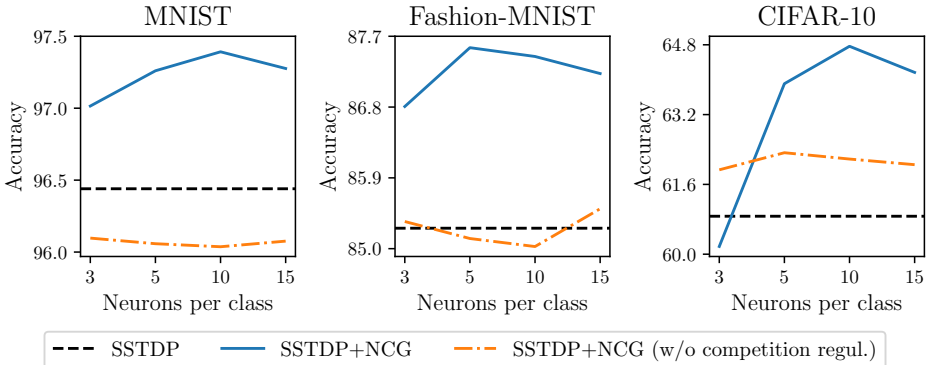

Figure 3: Accuracy against the number of neurons per class in the classification layer trained with SSTDP+NCG. The features are extracted with STDP-CSNN.

We analyze the impact of varying the number of neurons per class ($M$) on the performance of NCGs. Figures 2 and 3 present the accuracy achieved by S2-STDP+NCG and SSTDP+NCG using different numbers of neurons, with and without competition regulation. First, competition regulation is crucial to maximize performance with NCGs, especially for complex datasets. With competition regulation, increasing the number of neurons per class does not always result in better performance. For simpler datasets (MNIST, Fashion-MNIST), this increase often leads to either a plateau in accuracy or a decline when $M > 5$. Conversely, for harder datasets (CIFAR-10), using larger values ($M = 10$) improves the results from the main paper (which uses $M = 5$). Hence, the optimal number of neurons per class depends on the complexity of the task and intra-class pattern variations. It should be optimized as a hyperparameter. It is important to recall that the number of parameters in the

19

classification layer scales with the number of neurons. Hence, NCGs improve the performance of a classification layer at the cost of an increased number of parameters. Figures 2 and 3 show that significant accuracy gains can usually be achieved with only three neurons per class. This reveals that our method remains effective even with a minimal number of additional parameters. When selecting the value of $M$, the tradeoff between optimizing accuracy and minimizing parameter cost must be considered.

### 3.3 Impact of Hyperparameters

We study the impact of the threshold learning rate ($\eta_{th}$) and the threshold annealing ($\beta_{th}$), introduced by our competition regulation mechanism, on the accuracy. Figures 4 and 5 provide results for S2-STDP+NCG with STDP-CSNN and SoftHebb-CNN, respectively. Figures 6 and 7 provide results for SSTDP+NCG with STDP-CSNN and SoftHebb-CNN, respectively. Without annealing ($\beta_{th} = 1$), selecting an appropriate learning rate is crucial to obtain optimal performance. This is attributed to the increased difficulty in achieving convergence when employing higher learning rates that remain fixed during the entire training process. With annealing ($\beta_{th} < 1$), the performance of S2-STDP+NCG tends to be near-optimal regardless of the selected learning rate and annealing values. SSTDP+NCG is less robust to hyperparameters and requires more tuning to achieve optimal performance. This behavior may occur because SSTDP without NCG also exhibits lower hyperparameter robustness compared to S2-STDP [4]. However, the accuracy of SSTDP is still improved regardless of the hyperparameter values (when $\beta_{th} < 1$). Overall, annealing the threshold learning rate is not essential for benefiting from balanced competition, but it does offer better robustness against the selected value and can further improve performance.

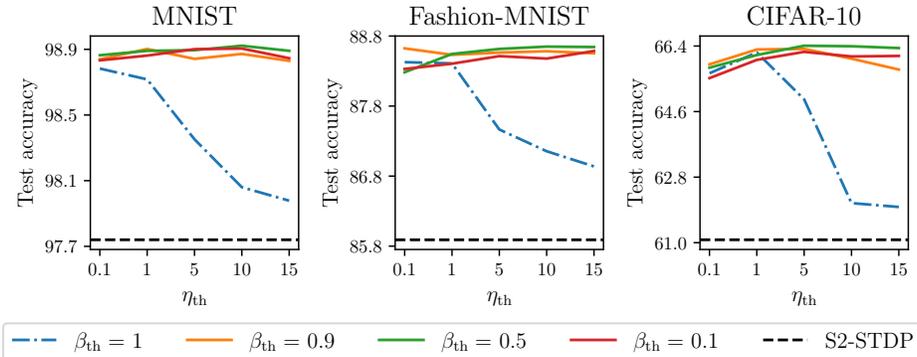

Figure 4: Accuracy of S2-STDP+NCG with different threshold annealing $\beta_{th}$ against the threshold learning rate $\eta_{th}$. The features are extracted with STDP-CSNN.

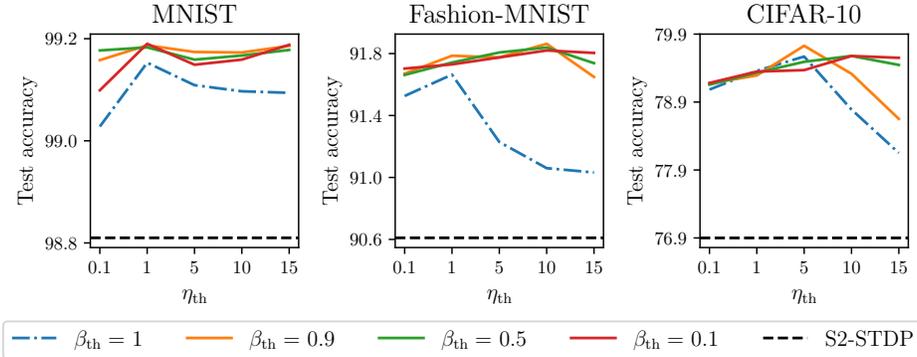

Figure 5: Accuracy of S2-STDP+NCG with different threshold annealing $\beta_{th}$ against the threshold learning rate $\eta_{th}$. The features are extracted with SoftHebb-CNN.

20

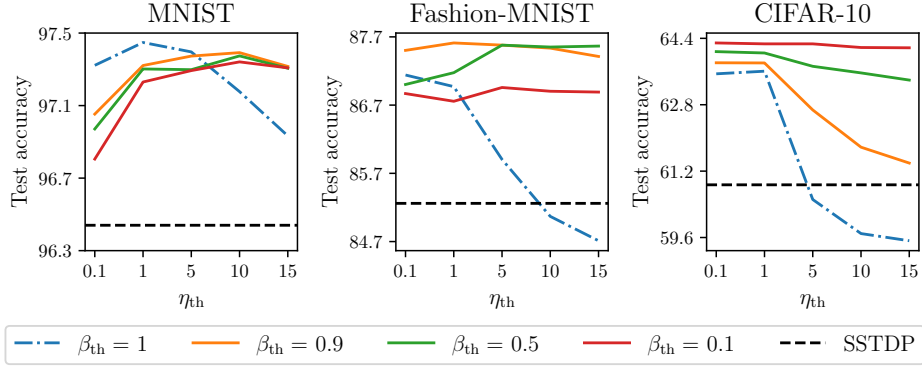

Figure 6: Accuracy of SSTDP+NCG with different threshold annealing $\beta_{\text{th}}$ against the threshold learning rate $\eta_{\text{th}}$. The features are extracted with STDP-CSNN.

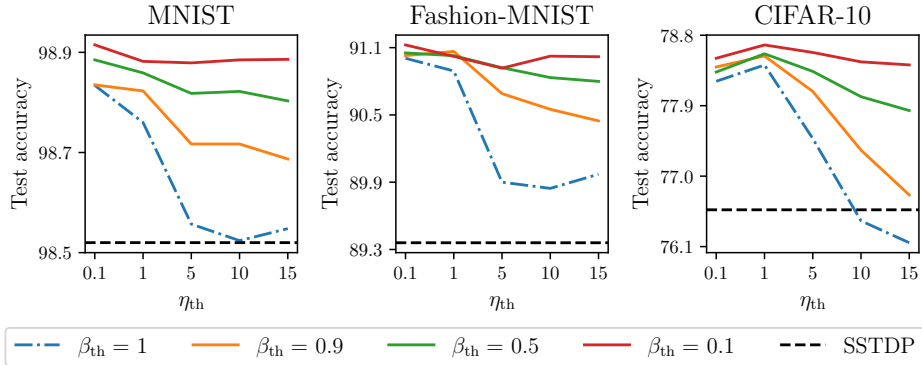

Figure 7: Accuracy of SSTDP+NCG with different threshold annealing $\beta_{\text{th}}$ against the threshold learning rate $\eta_{\text{th}}$. The features are extracted with SoftHebb-CNN.

### 3.4 Ablation Study

Table 2: Ablation study on SSTDP+NCG. *M* is the number of neurons per class, *CR* is competition regulation with 1 or 2 thresholds, and *Drop* is dropout.

(a) Fashion-MNIST

| Method | Accuracy (Mean±Std %) STDP-CSNN | SoftHebb-CNN |
|---|---|---|
| *M-1* | $85.26 \pm 0.17$ | $89.36 \pm 0.24$ |
| *M-5* | $85.13 \pm 1.20$ | $91.02 \pm 0.14$ |
| *M-5+CR-1* | $86.79 \pm 0.14$ | $90.92 \pm 0.22$ |
| *M-5+CR-2* | $\mathbf{87.59 \pm 0.11}$ | $\mathbf{91.06 \pm 0.10}$ |
| *M-5+Drop* | $86.40 \pm 0.27$ | $90.90 \pm 0.14$ |

(b) CIFAR-10

| Method | Accuracy (Mean±Std %) STDP-CSNN | SoftHebb-CNN |
|---|---|---|
| *M-1* | $60.87 \pm 0.53$ | $76.57 \pm 0.58$ |
| *M-5* | $62.33 \pm 0.14$ | $77.51 \pm 0.23$ |
| *M-5+CR-1* | $63.50 \pm 0.28$ | $78.09 \pm 0.34$ |
| *M-5+CR-2* | $\mathbf{64.05 \pm 0.48}$ | $\mathbf{78.53 \pm 0.32}$ |
| *M-5+Drop* | $62.46 \pm 0.19$ | $78.12 \pm 0.22$ |

In the main paper, we conduct an ablation study on S2-STDP+NCG to show the individual accuracy gain brought by each component of our methods. We provide, in Table 2, a similar study on SSTDP+NCG. *M-1* and *M-5* represent SSTDP+NCG with $M = 1$ (one neuron per class, which is similar to SSTDP) and $M = 5$, without competition regulation. *CR-1* denotes our competition regulation mechanism with a single threshold per neuron (i.e. $\theta' = \theta$). Thresholds are not clipped nor reset between epochs, and the learned values are used for inference. *CR-2* denotes our competition regulation with two-compartment thresholds. *Drop* is dropout on the output neurons (an alternative competition regulation mechanism used with R-STDP). Neuron labeling is not employed because

21

SSTDP uses error clipping, and neuron specialization is only relevant without error clipping [4]. Overall, these results are consistent with the study conducted on S2-STDP+NCG. Competition regulation improves class separation (cf. *M-5*), except on Fashion-MNIST with SoftHebb-CNN, where the accuracy improvement is solely attributed to intra-class WTA. Our competition regulation mechanism (cf. *CR-2*) outperforms the existing threshold adaptation with one threshold (cf. *CR-1*), as well as dropout (cf. *Drop*).

### 3.5 Impact of Competition Regulation

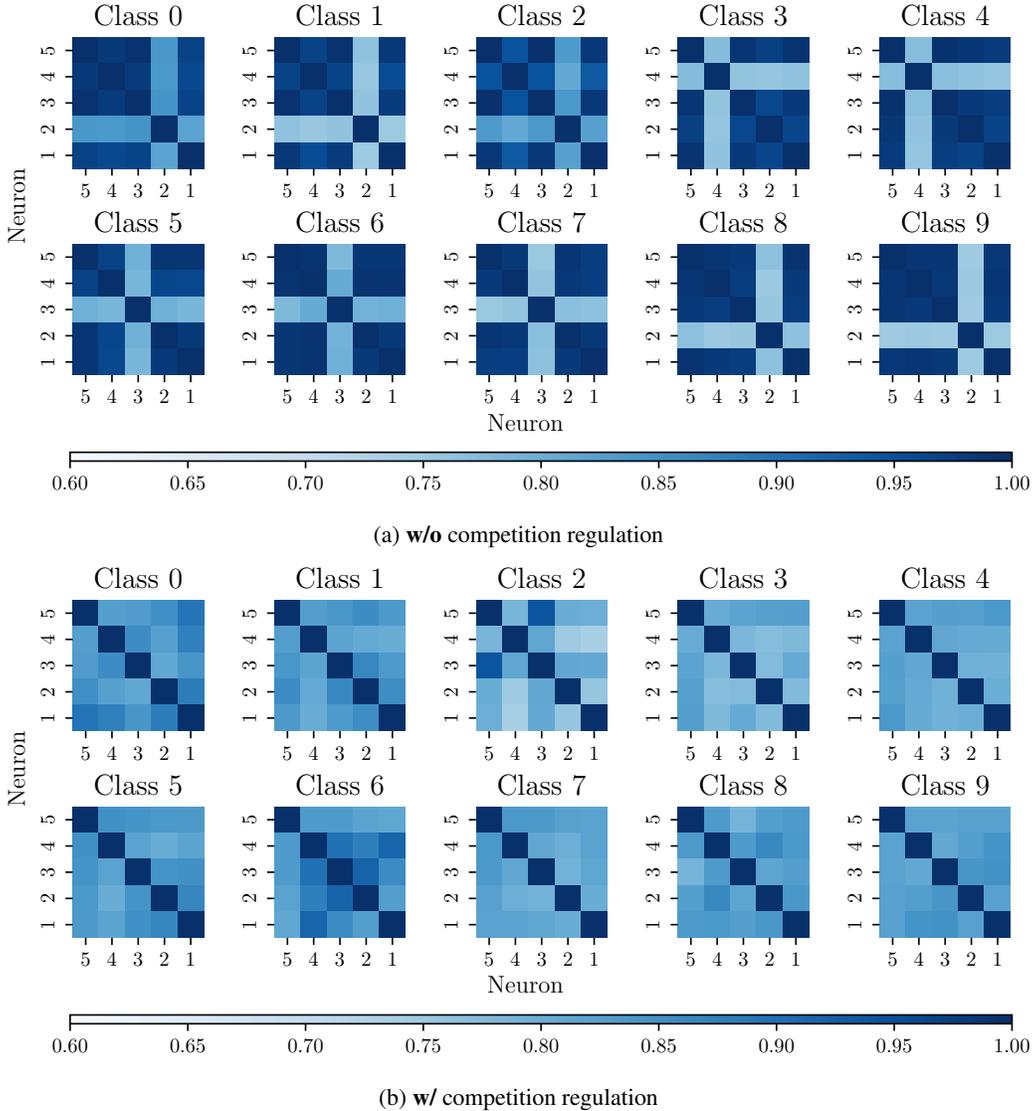

Figure 8: Heatmap of intra-class weight cosine similarities, after S2-STDP+NCG training on CIFAR-10. The features are extracted with STDP-CSNN.

In the main paper, we suggest through t-SNE visualizations that competition regulation enables the learning of various patterns per class. In this section, we further analyze the weights learned with and without competition regulation to support this claim. Figure 8 illustrates intra-class weight cosine similarities, after S2-STDP+NCG training on CIFAR-10. Without competition regulation, the neuron with the lowest score (such as neuron 2 of class 0) is the neuron receiving the majority of the target updates during training (see Figure 9a). All the other neurons exhibit high scores, indicating significant similarity in their weights. Competition regulation successfully decreases the similarity

22

score between neurons of each NCG (i.e. class), which provides further evidence that it enables the learning of various class-specific patterns.

We present additional figures to show that competition regulation is crucial for ensuring balanced competition. Figures 9 and 10 illustrate the number of target weight updates received by neurons trained with S2-STDP+NCG on CIFAR-10 and Fashion-MNIST, respectively. Figures 11 and 12 present similar plots with SSTDP+NCG. Without competition regulation, one neuron of each NCG (i.e. class) receives the majority of the target updates. Our competition regulation mechanism successfully balances the number of updates received by neurons, which promotes, through intra-class WTA, the learning of various patterns per class. Note that competition for some classes is not completely balanced, particularly on Fashion-MNIST. However, absolute balance in the competition is not expected, given that the training set does not necessarily ensure equal representation of all different patterns.

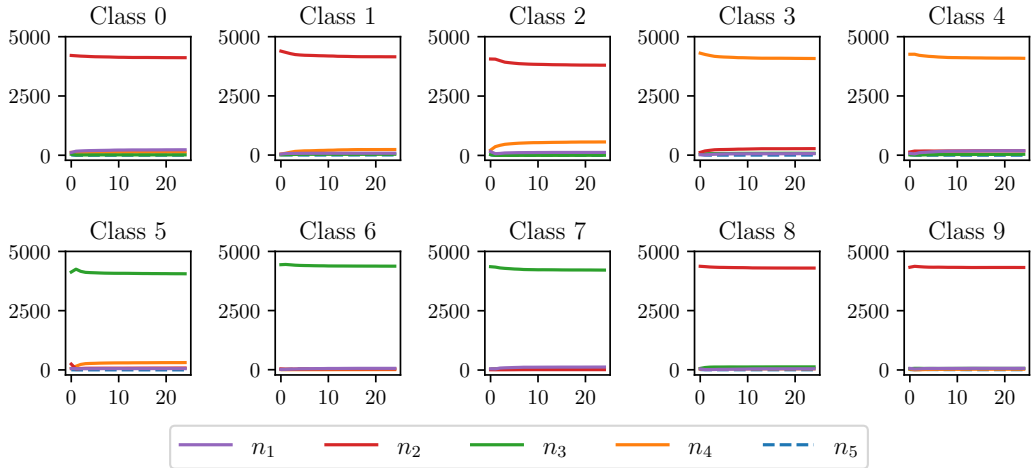

(a) **w/o** competition regulation

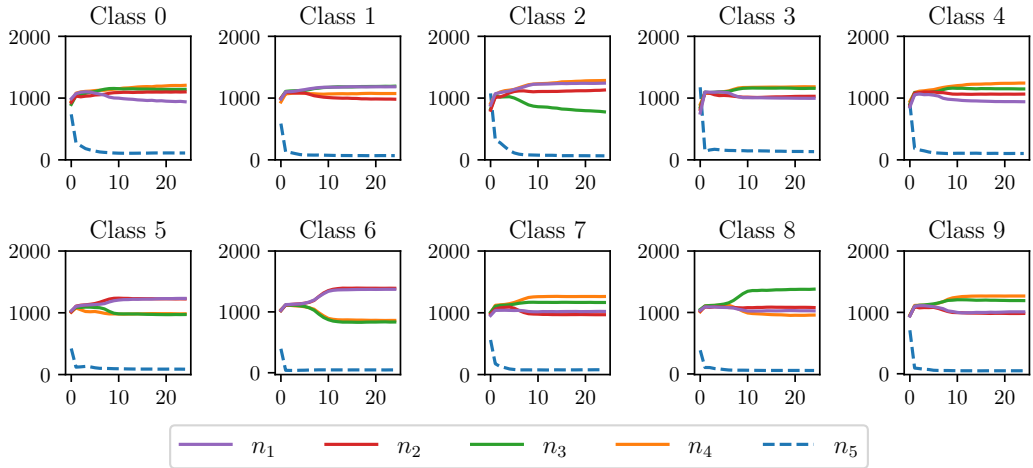

(b) **w/** competition regulation

Figure 9: Number of target weight updates per epoch received by neurons of each class on CIFAR-10, with S2-STDP+NCG training. $n_1$ to $n_4$ are labeled as target neurons and $n_5$ is labeled as non-target. The features are extracted with STDP-CSNN.

23

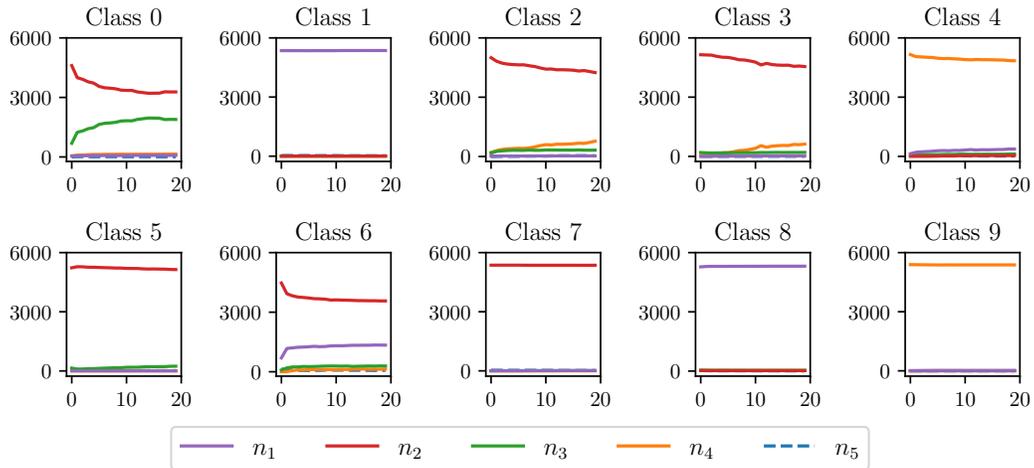

(a) **w/o** competition regulation

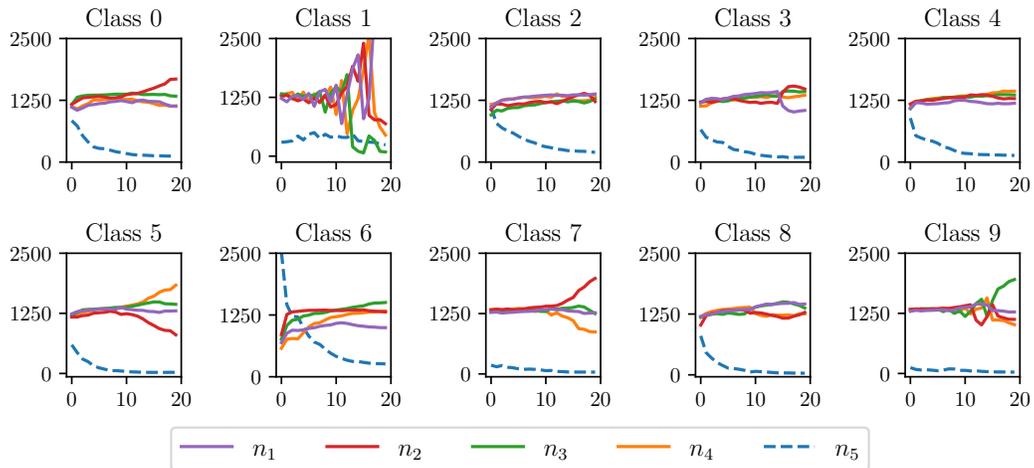

(b) **w/** competition regulation

Figure 10: Number of target weight updates per epoch received by neurons of each class on Fashion-MNIST, with S2-STDP+NCG training. $n_1$ to $n_4$ are labeled as target neurons and $n_5$ is labeled as non-target. The features are extracted with STDP-CSNN.

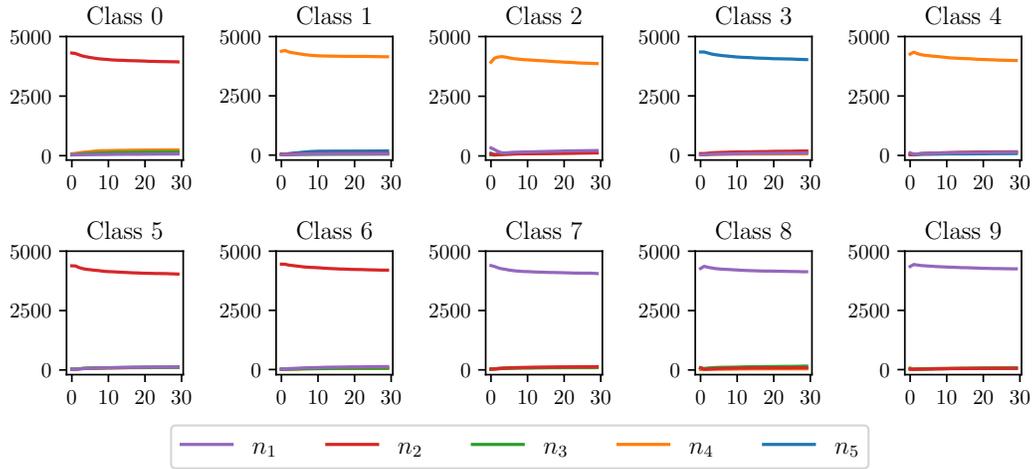

(a) **w/o** competition regulation

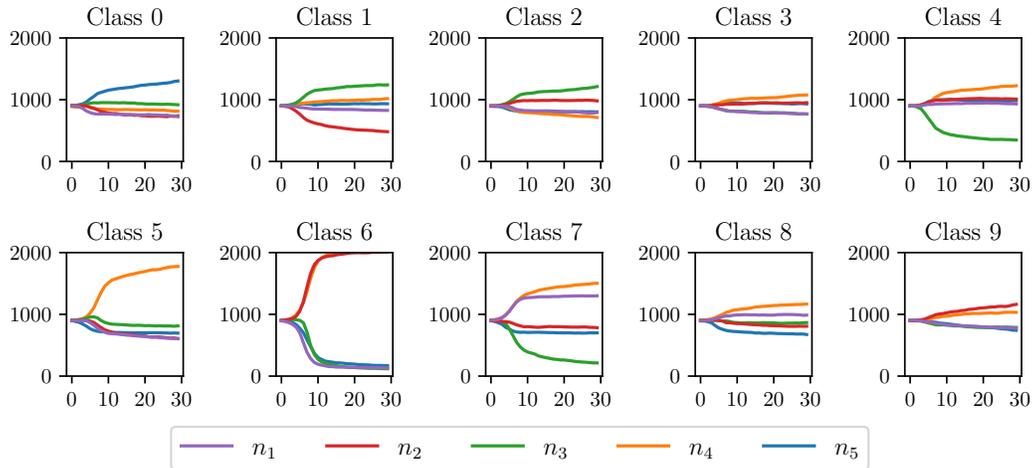

(b) **w/** competition regulation

Figure 11: Number of target weight updates per epoch received by neurons of each class on CIFAR-10, with SSTDP+NCG training. The features are extracted with STDP-CSNN.

25

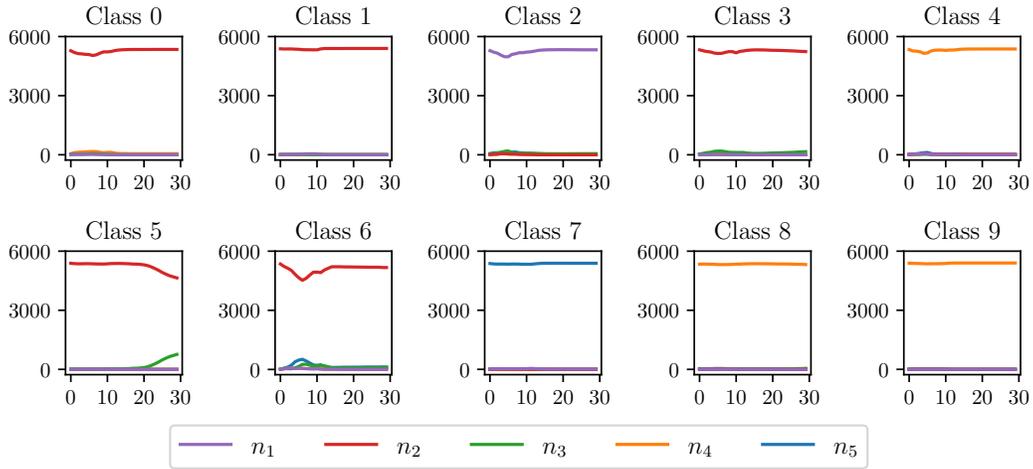

(a) **w/o** competition regulation

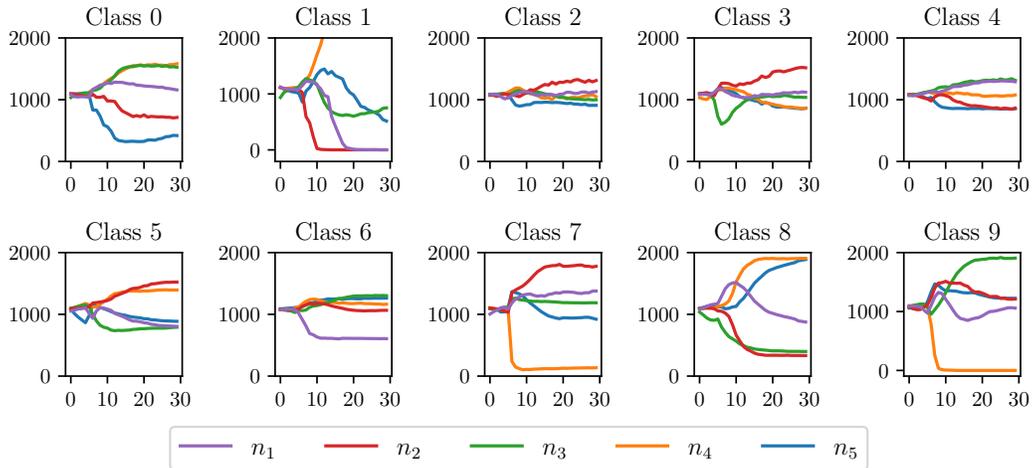

(b) **w/** competition regulation

Figure 12: Number of target weight updates per epoch received by neurons of each class on Fashion-MNIST, with SSTDP+NCG training. The features are extracted with STDP-CSNN.

## 3.6 Application to Gradient-Based Learning Rules

Table 3: Accuracy of S2-STDP+NCG, with and without long-term depression.

| Dataset | Long-term depression | Accuracy (Mean±Std %) STDP-CSNN | SoftHebb-CNN |
|---|---|---|---|
| MNIST | no | 98.71 ± 0.14 | 99.15 ± 0.05 |
|  | yes | **98.92 ± 0.07** | **99.17 ± 0.07** |
| Fashion-MNIST | no | 88.36 ± 0.12 | 91.54 ± 0.21 |
|  | yes | **88.72 ± 0.23** | **91.86 ± 0.14** |
| CIFAR-10 | no | 65.91 ± 0.38 | **79.57 ± 0.24** |
|  | yes | **66.41 ± 0.17** | 79.55 ± 0.23 |

To train the spiking classification layers, we employ an error-modulated additive STDP in which the weight change is the product of the error and the learning rate (see Equation 3, page 4, in the main paper). The learning rate is positive for long-term potentiation (i.e. when the input neuron fires before the output neuron) and negative for long-term depression (i.e. when the input neuron fires after the output neuron). Given the simplicity of this STDP model, the weight updates, if long-term depression is ignored, resemble a gradient-based rule or delta rule[1]. To better understand the relevance of STDP with respect to these types of rules, we examine, in Table 3, the accuracy of S2-STDP+NCG with and without long-term depression. Results indicate that incorporating long-term depression generally leads to a slight improvement in accuracy. This improvement may be due to STDP considering input spikes that reach the neuron both before and after the output spike. In addition, long-term depression enables faster training convergence by increasing the number of weight updates per sample. The number of epochs with long-term depression is reduced by an average of 15% for STDP-CSNN and 4% for SoftHebb-CNN. As a result, S2-STDP enables more effective training of the NCGs than a gradient-based rule that uses a squared error loss function and the same method for defining the desired firing times. However, it should be noted that the method for computing the errors (specifically, the desired firing times) is more important than the method for updating the weights (gradient-based or STDP-based).

Table 4: Accuracy of spiking classification layers trained with a gradient-based method, on top of Hebbian-based unsupervised feature extractors.

| Dataset | Method | Neurons per class | Accuracy (Mean±Std %) STDP-CSNN | SoftHebb-CNN |
|---|---|---|---|---|
| MNIST | S4NN | 1 | 97.86 ± 0.11 | 98.86 ± 0.14 |
|  | S4NN+NCG (ours) | 5 | **98.33 ± 0.10** | **98.99 ± 0.09** |
| Fashion-MNIST | S4NN | 1 | 86.55 ± 0.13 | 90.38 ± 0.32 |
|  | S4NN+NCG (ours) | 5 | **87.87 ± 0.24** | **90.98 ± 0.14** |
| CIFAR-10 | S4NN | 1 | 57.86 ± 0.15 | 76.92 ± 0.26 |
|  | S4NN+NCG (ours) | 5 | **62.55 ± 0.20** | **77.62 ± 0.27** |
| CIFAR-100 | S4NN | 1 | 24.38 ± 0.44 | 38.51 ± 1.00 |
|  | S4NN+NCG (ours) | 5 | **28.68 ± 0.61** | **39.84 ± 0.54** |

To provide additional evidence that NCGs can be trained with gradient-based rules, we conducted another experiment using S4NN [9], an established rule for single-spike neurons. S4NN uses the stochastic gradient descent algorithm (with gradient approximations) to minimize a squared error loss function, and computes the desired firing times based on the first output firing time. Table 4 presents the accuracy achieved by this rule, with and without NCGs, across various datasets and

---
[1] The delta learning rule must employ the same method for computing the errors and a Heaviside function to convert input spikes into a continuous signal.

feature extractors. NCGs consistently improve the performance of S4NN, aligning with the results from the main paper on STDP-based rules. Yet, further research is required to evaluate the impact of the method for computing the errors (and the desired firing times) on the effectiveness of NCGs.

## 4 Comparison with SOTA Methods

SOTA methods for direct training of SNNs [10, 11, 12] rely on backpropagation through time (BPTT) [13] and surrogate gradient [14]. These methods usually allow multiple spikes per neuron and support the training of very deep networks using global supervised learning. In this work, we allow one spike per neuron, train all layers with local learning (limiting our networks to shallow architectures), and use a semi-supervised training strategy, where only the last layer is trained with supervision. In terms of performance, our methods lag behind fully-supervised SOTA methods. For instance, [12] report an accuracy of $96.44\%$ on CIFAR-10 (our best model achieves $79.55\%$) and $81.65\%$ on CIFAR-100 (our best model achieves $53.49\%$). This decrease in accuracy can partially be attributed to the number of layers employed (4 against 19) and the use of supervision limited to the last layer. In terms of computational and memory costs, BPTT is extremely inefficient since these costs scale with the latency (i.e. the number of time steps), whereas the costs of our methods are independent of the latency. Also, a backward pass with BPTT adjusts all synapses in the network, whereas our methods adjust only the synapses of neurons that have fired. In terms of energy efficiency, our single-spike strategy may limit the number of generated spikes significantly compared to multiple-spike methods, which reduces power consumption in both training and inference. In terms of hardware suitability, BPTT is challenging to implement on neuromorphic hardware because it relies on non-local learning [15]. BPTT-based SNNs must be trained on GPUs, which is energy-intensive [16, 17], and can be deployed on chip for inference only. To fully exploit the energy-efficient capabilities of SNNs, both training and inference should be performed on chip. We target, for instance, memristive-based chips [18] for hardware implementation of our methods. They are excellent candidates for ultra-low-power applications, potentially reducing energy consumption by several orders of magnitude compared to GPUs [19, 16]. Also, STDP is inherently implemented in memristor circuits [20, 21], which facilitates on-chip training [22, 23]. There are still several challenges to address before our work can be implemented on this type of chip, such as the need for a digital module to calculate the error. This should be the focus of future work.

## References

[1] Pierre Falez, Pierre Tirilly, Ioan Marius Bilasco, Philippe Devienne, and Pierre Boulet. Multi-Layered Spiking Neural Network with Target Timestamp Threshold Adaptation and STDP. In *International Joint Conference on Neural Networks*, 2019.

[2] Pierre Falez, Pierre Tirilly, and Ioan Marius Bilasco. Improving STDP-based Visual Feature Learning with Whitening. In *International Joint Conference on Neural Networks*, 2020.

[3] Simon Thorpe, Arnaud Delorme, and Rufin Van Rullen. Spike-Based Strategies for Rapid Processing. *Neural Networks*, 14:715–725, 2001.

[4] Gaspard Goupy, Pierre Tirilly, and Ioan Marius Bilasco. Paired Competing Neurons Improving STDP Supervised Local Learning in Spiking Neural Networks. *Frontiers in Neuroscience*, 18, 2024.

[5] Adam Coates, Andrew Ng, and Honglak Lee. An Analysis of Single-Layer Networks in Unsupervised Feature Learning. In *International Conference on Artificial Intelligence and Statistics*, pages 215–223, 2011.

[6] Adrien Journé, Hector Garcia Rodriguez, Qinghai Guo, and Timoleon Moraitis. Hebbian Deep Learning Without Feedback. *International Conference on Learning Representations*, 2023.

[7] Milad Mozafari, Saeed Reza Kheradpisheh, Timothee Masquelier, Abbas Nowzari-Dalini, and Mohammad Ganjtabesh. First-Spike-Based Visual Categorization Using Reward-Modulated STDP. *Transactions on Neural Networks and Learning Systems*, 29:6178–6190, 2018.

[8] Franck Cappello, Frédéric Desprez, Michel Daydé, Emmanuel Jeannot, Yvon Jégou, Stephane Lanteri, Nouredine Melab, Raymond Namyst, Pascale Primet, Olivier Richard, Eddy Caron, Julien Leduc, and Guillaume Mornet. Grid'5000: A Large Scale, Reconfigurable, Controlable and Monitorable Grid Platform. In *International Workshop on Grid Computing*, 2005.

[9] Saeed Reza Kheradpisheh and Timothée Masquelier. Temporal Backpropagation for Spiking Neural Networks with One Spike per Neuron. *International Journal of Neural Systems*, 30, 2020.

[10] Chaoteng Duan, Jianhao Ding, Shiyan Chen, Zhaofei Yu, and Tiejun Huang. Temporal Effective Batch Normalization in Spiking Neural Networks. *Advances in Neural Information Processing Systems*, 35, 2022.

[11] Man Yao, JiaKui Hu, Zhaokun Zhou, Li Yuan, Yonghong Tian, Bo Xu, and Guoqi Li. Spike-driven Transformer. *Advances in Neural Information Processing Systems*, 36, 2023.

[12] Yuhang Li, Tamar Geller, Youngeun Kim, and Priyadarshini Panda. SEENN: Towards Temporal Spiking Early Exit Neural Networks. *Advances in Neural Information Processing Systems*, 36, 2023.

[13] Yujie Wu, Lei Deng, Guoqi Li, Jun Zhu, and Luping Shi. Spatio-Temporal Backpropagation for Training High-Performance Spiking Neural Networks. *Frontiers in Neuroscience*, 12, 2018.

[14] Emre O. Neftci, Hesham Mostafa, and Friedemann Zenke. Surrogate Gradient Learning in Spiking Neural Networks: Bringing the Power of Gradient-Based Optimization to Spiking Neural Networks. *Signal Processing Magazine*, 36:51–63, 2019.

[15] Friedemann Zenke and Emre Neftci. Brain-Inspired Learning on Neuromorphic Substrates. *Proceedings of the IEEE*, 109:935–950, 2021.

[16] Jiwei Li, Hui Xu, Sheng-Yang Sun, Nan Li, Qingjiang Li, Zhiwei Li, and Haijun Liu. In Situ Learning in Hardware Compatible Multilayer Memristive Spiking Neural Network. *Transactions on Cognitive and Developmental Systems*, 14:448–461, 2022.

[17] Shiya Liu, Nima Mohammadi, and Yang Yi. Quantization-Aware Training of Spiking Neural Networks for Energy-Efficient Spectrum Sensing on Loihi Chip. *Transactions on Green Communications and Networking*, 2023.

[18] Doo Seok Jeong, Kyung Min Kim, Sungho Kim, Byung Joon Choi, and Cheol Seong Hwang. Memristors for Energy-Efficient New Computing Paradigms. *Advanced Electronic Materials*, 2, 2016.

[19] Peng Yao, Huaqiang Wu, Bin Gao, Jianshi Tang, Qingtian Zhang, Wenqiang Zhang, J. Joshua Yang, and He Qian. Fully Hardware-Implemented Memristor Convolutional Neural Network. *Nature*, 577:641–646, 2020.

[20] Damien Querlioz, Olivier Bichler, and Christian Gamrat. Simulation of a Memristor-Based Spiking Neural Network Immune to Device Variations. In *International Joint Conference on Neural Networks*, pages 1775–1781, 2011.

[21] Catherine D. Schuman, Thomas E. Potok, Robert M. Patton, J. Douglas Birdwell, Mark E. Dean, Garrett S. Rose, and James S. Plank. A Survey of Neuromorphic Computing and Neural Networks in Hardware. *ArXiv*, arXiv:1705.06963 [cs.NE], 2017.

[22] Sylvain Saïghi, Christian G. Mayr, Teresa Serrano-Gotarredona, Heidemarie Schmidt, Gwendal Lecerf, Jean Tomas, Julie Grollier, Sören Boyn, Adrien F. Vincent, Damien Querlioz, Selina La Barbera, Fabien Alibart, Dominique Vuillaume, Olivier Bichler, Christian Gamrat, and Bernabé Linares-Barranco. Plasticity in Memristive Devices for Spiking Neural Networks. *Frontiers in Neuroscience*, 9, 2015.

[23] Lyes Khacef, Philipp Klein, Matteo Cartiglia, Arianna Rubino, Giacomo Indiveri, and Elisabetta Chicca. Spike-Based Local Synaptic Plasticity: A Survey of Computational Models and Neuromorphic Circuits. *Neuromorphic Computing and Engineering*, 3, 2023.



\end{document}